\documentclass[10pt,twocolumn,letterpaper]{article}

\PassOptionsToPackage{table}{xcolor}

\usepackage[pagenumbers]{cvpr}

\definecolor{cvprblue}{rgb}{0.21,0.49,0.74}
\usepackage[pagebackref,breaklinks,colorlinks,allcolors=cvprblue]{hyperref}
\usepackage{amsmath}
\usepackage{graphicx}
\usepackage{booktabs}
\usepackage{multirow}
\usepackage{colortbl}
\usepackage{makecell}
\usepackage{float}
\usepackage{placeins}

\def\paperID{*****} 
\def\confName{CVPR}
\def\confYear{2026}

\title{PLUME: Latent Reasoning Based Universal Multimodal Embedding}

\author{
Chenwei He$^{1}$\thanks{Equal contribution.} \quad
Xiangzhao Hao$^{2,3}$\footnotemark[1] \quad
Tianyu Yang$^{2,3}$\footnotemark[1] \quad
Yuxiang Ma$^{1}$ \quad
Yuheng Jia$^{1}$ \\
Lingxiang Wu$^{2,3}$ \quad
Chaoyang Zhao$^{2,3}$ \quad
Haiyun Guo$^{2,3}$\thanks{Corresponding author.} \quad
Jinqiao Wang$^{2,3}$ \\
$^{1}$Southeast University \\
$^{2}$Institute of Automation, Chinese Academy of Sciences \\
$^{3}$University of Chinese Academy of Sciences \\
{\tt\small \{hechenwei, 220256453, yhjia\}@seu.edu.cn} \\
{\tt\small \{haoxiangzhao2023, yangtianyu2024\}@ia.ac.cn} \\
{\tt\small \{lingxiang.wu, chaoyang.zhao, haiyun.guo, jqwang\}@nlpr.ia.ac.cn}
}

\begin{document}
\maketitle

\begin{abstract}
Universal multimodal embedding (UME) maps heterogeneous inputs into a shared retrieval space with a single model. Recent approaches improve UME by generating explicit chain-of-thought (CoT) rationales before extracting embeddings, enabling multimodal large language models to better infer complex query intent. However, explicit CoT incurs substantial inference overhead and can compress rich multimodal evidence into a narrow textual bottleneck. We propose PLUME, a latent reasoning framework that advances UME by replacing verbalized CoT with a short autoregressive rollout of continuous latent states. To support diverse multimodal queries, PLUME further introduces a semantic-anchor-guided transition adapter that steers latent rollout along different reasoning trajectories under the same fixed computation budget. To stabilize training, PLUME adopts a progressive explicit-to-latent curriculum that uses verbalized reasoning only as a temporary training scaffold and gradually transfers this behavior into hidden-state computation, eliminating explicit CoT at inference. On the 78-task MMEB-v2 benchmark, PLUME scores 61.6 at 2B and 67.1 at 7B, exceeding the size-matched explicit-CoT baseline UME-R1 by 1.5 and 2.6 points, respectively. At 2B, eight latent steps replace an average of 403 generated reasoning tokens and yield a $30.3\times$ inference speedup. PLUME is especially well suited to retrieval settings where relevant evidence is dense, structurally complex, and difficult to organize through verbalized intermediate rationales, such as video and visual document retrieval. These results show that structured latent computation can preserve the benefits of intermediate reasoning without the overhead of explicit rationale generation, providing a stronger and more efficient paradigm for practical retrieval systems. Our code and data are publicly available at \url{https://github.com/haoxiangzhao12138/PLUME}.

\end{abstract}

\section{Introduction}

\begin{figure}[t]
    \centering
    \includegraphics[width=\linewidth]{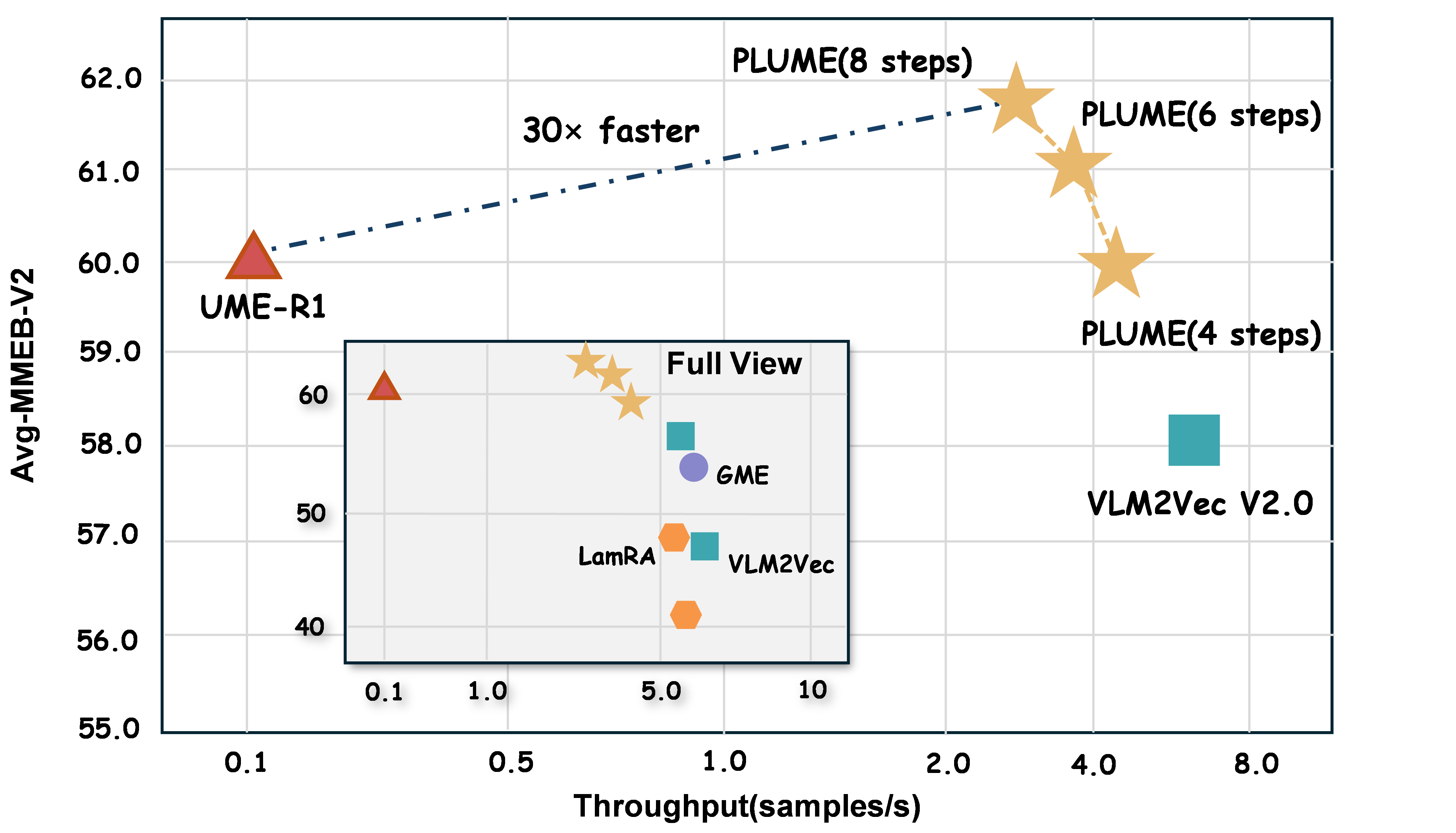}
    \caption{\textbf{PLUME achieves a favorable accuracy--efficiency tradeoff on MMEB-v2.} For the 2B models, the x-axis shows inference throughput on a single H20 GPU and the y-axis shows average MMEB-v2 performance.}
    \label{fig:pareto}
\end{figure}

\begin{figure*}[t]
    \centering
    \includegraphics[width=\textwidth]{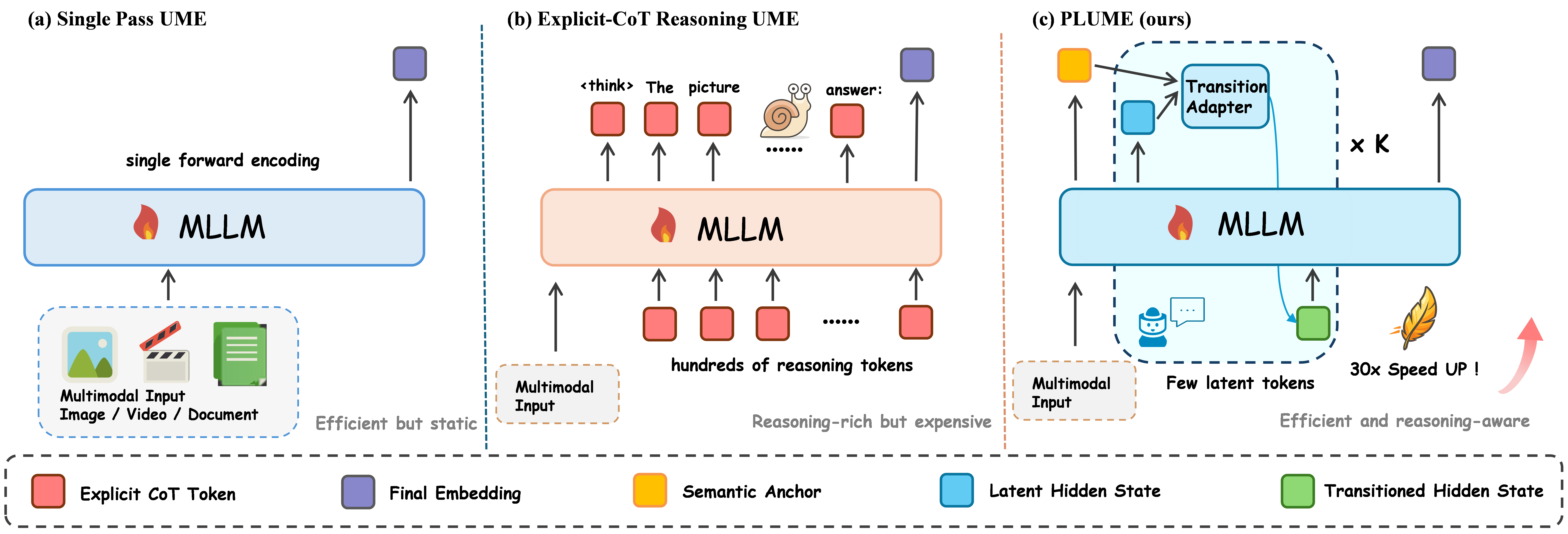}
    \caption{Comparison of three universal multimodal embedding paradigms. Left: early discriminative UME forms embeddings through single pass encoding, preserving efficiency but without explicitly modeling intermediate reasoning. Middle: explicit CoT UME improves reasoning by generating long textual traces before embedding extraction, but incurs substantial inference latency and token cost. Right: PLUME internalizes reasoning into a compact latent rollout and adapts the reasoning path with semantic-anchor-guided expert routing, achieving reasoning-aware embedding with substantially lower inference cost.}
    \label{fig:intro_comparison}
\end{figure*}

Universal multimodal embedding (UME) aims to map heterogeneous inputs, including text, images, videos, and visual documents, into a shared retrieval space with a single model \cite{jiang2024vlm2vec,zhang2024gme,meng2025vlm2vec}. In real-world retrieval, however, many queries cannot be resolved by surface-level similarity alone. They often require compositional spatial understanding, knowledge-intensive visual inference, or the aggregation of temporally and structurally dispersed evidence. These demands have made Multimodal Large Language Models (MLLMs) \cite{liu2023visual,li2024llava,wang2024qwen2,grattafiori2024llama} an increasingly attractive backbone for UME, thanks to their native multimodal grounding, strong semantic alignment, and broad world knowledge. Yet simply adopting an MLLM as the encoder does not automatically translate its reasoning potential into stronger embeddings\cite{yang2026recall}. In most existing UME pipelines, the embedding is still formed in a single pass, leaving limited room for deliberate intermediate computation when query intent is complex. This raises a central question for UME: how can we leverage the reasoning capability of MLLMs during embedding formation without sacrificing retrieval efficiency?

Existing attempts to address this question mainly follow two directions, as illustrated in Figure~\ref{fig:intro_comparison}. Single-pass MLLM-based methods \cite{jiang2024e5,jiang2024vlm2vec,lin2024mm,zhang2024gme} are efficient, but they require the model to collapse complex query interpretation, evidence integration, and representation formation into one forward pass. To better handle such complexity, recent reasoning-enhanced methods, such as TTE \cite{cui2025think}, UME-R1 \cite{lan2025ume}, and TRACE \cite{hao2026tracetaskadaptivereasoningrepresentation}, first generate an explicit chain-of-thought (CoT) rationale \cite{wei2022chain,wang2022self} before deriving the final embedding. While effective, this strategy introduces a dual bottleneck. Computationally, generating hundreds of reasoning tokens per sample incurs substantial autoregressive decoding overhead and severely limits inference throughput. Representationally, routing multimodal reasoning through discrete textual tokens creates a narrow bottleneck that may discard fine-grained continuous evidence and constrain how richly multimodal information is carried into the final embedding. As a result, explicit CoT ties the benefits of multi-step computation to a verbose interface that is fundamentally mismatched with the efficiency demands of retrieval.

In light of this, we take a different perspective on UME: what retrieval needs is intermediate computation, not necessarily verbalized intermediate text. As illustrated in Figure~\ref{fig:intro_comparison}(c), the multi-step reasoning that helps embedding quality can instead unfold directly in the continuous hidden space of the backbone \cite{coconut,codi,latent_cot_survey}. A short latent rollout can preserve the sequential dependency structure of reasoning while avoiding long-form text generation. Yet moving from explicit reasoning to latent reasoning is not a trivial substitution in the multimodal setting. Unlike pure language tasks, UME must handle videos, images, documents, and text within one shared framework, and these inputs demand different forms of intermediate computation over temporal dynamics, spatial relations, layout structure, and semantic abstraction. Once reasoning is executed within a short latent budget, the key challenge is no longer whether to reason, but how to allocate this compact latent computation adaptively across heterogeneous multimodal queries instead of forcing every input through the same fixed reasoning path.

To tackle above issues, we propose \textbf{PLUME}, a latent reasoning framework for universal multimodal embedding. PLUME internalizes reasoning for UME into a compact latent process inside the MLLM, allowing the model to perform multi-step computation without generating explicit rationales. To make this latent process suitable for structurally diverse multimodal inputs, PLUME further introduces a \textbf{semantic-anchor-guided transition adapter} that steers latent computation according to the semantic structure of the input, enabling different queries to follow different reasoning patterns under the same rollout budget. Finally, rather than viewing explicit CoT merely as a costly inference procedure, PLUME uses it as a temporary training scaffold. During training, the model is first exposed to verbalized intermediate reasoning and then gradually shifts reasoning process into latent rollouts, progressively replacing explicit textual rationales with hidden-state computation until explicit CoT is no longer needed at inference time.

Experiments on MMEB-v2 show that PLUME outperforms UME-R1 on both Qwen2-VL-2B and Qwen2-VL-7B. With the 2B backbone, PLUME uses eight latent steps and runs $30.3\times$ faster than UME-R1 (Figure~\ref{fig:pareto}). PLUME is particularly effective on retrieval tasks where relevant evidence is dense, structurally complex, and difficult to organize through verbalized intermediate rationales, such as video and visual document retrieval. Taken together, these results suggest that strong UME benefits more from adaptive intermediate computation than from explicit verbalized rationales. By retaining reasoning quality inside a compact latent process, PLUME breaks the dual bottleneck of explicit CoT, bringing the benefits of intermediate reasoning back into the efficiency regime required by practical UME systems.

\smallskip\noindent
In summary, our contributions are:

\begin{itemize}

\item \textbf{A latent reasoning framework for UME.}
We introduce PLUME to internalize intermediate reasoning into a short continuous latent process for UME, replacing costly explicit chain-of-thought generation while preserving the benefits of intermediate computation.

\item \textbf{An input-adaptive latent reasoning architecture.}
We design a semantic-anchor-guided transition adapter that allocates latent computation adaptively across heterogeneous multimodal queries, allowing the same compact rollout budget to support different reasoning patterns for images, videos, documents, and text.

\item \textbf{Strong empirical gains in both effectiveness and efficiency.}
PLUME improves video and visual document retrieval over UME-R1 at both model sizes. The largest gains are on visual documents (+3.6 points at 2B and +5.3 at 7B). At 2B, PLUME also reduces per-sample inference latency from 9023\,ms to 298\,ms.
\end{itemize}

\noindent

\section{Related Work}

\subsection{Universal Multimodal Embedding}

Universal multimodal embedding (UME) maps heterogeneous inputs into a shared retrieval space with a single model. Early dual-encoder methods such as CLIP \cite{radford2021learning}, ALIGN \cite{jia2021scaling}, SigLIP \cite{zhai2023sigmoid}, and BLIP-2 \cite{li2023blip} learn aligned image-text representations via contrastive objectives \cite{oord2018representation} but are less effective on complex multimodal compositions. UniIR \cite{wei2024uniir} and MagicLens \cite{zhang2024magiclens} begin to address multi-task multimodal retrieval within unified frameworks. Building on advances in LLM-based text embedding \cite{e5,e5mistral,nvembed,bge}, MLLM-based approaches \cite{yang2026recallrecalibratingcapabilitydegradation} further overcome this limitation: E5-V \cite{jiang2024e5} and MM-Embed \cite{lin2024mm} prompt MLLMs for universal embeddings; VLM2Vec \cite{jiang2024vlm2vec} introduces the MMEB benchmark; and VLM2Vec-V2 \cite{meng2025vlm2vec}, GME \cite{zhang2024gme}, UniME \cite{gu2025breaking}, LamRA \cite{liu2025lamra}, LLaVE \cite{lan2025llave}, MoCa \cite{chen2025moca}, and the DUME non-CoT baseline reported in UME-R1 \cite{lan2025ume} further improve retrieval quality and modality coverage. More recent efforts explore multi-vector representations \cite{faysse2024colpali}, large-scale data synthesis \cite{zhou2025megapairs,zhou2024megapairs}, visual document retrieval \cite{yu2024visrag}, and reinforcement-learning-based alignment \cite{yu2025cafe} to push the accuracy--efficiency frontier. However most methods derive embeddings from a single forward pass without modeling intermediate reasoning, limiting performance on complex retrieval query.

\subsection{Reasoning-Enhanced Embedding}

Chain-of-thought (CoT) prompting \cite{wei2022chain,wang2022self} elicits multi-step reasoning in language models, and subsequent extensions such as multimodal CoT \cite{xu2025llava} and preference-optimized reasoning \cite{zhang2024chain} further strengthen reasoning quality. Scaling this idea, reasoning-specialized LLMs such as DeepSeek-R1 \cite{guo2025deepseek} have demonstrated the power of long-form reasoning. Recent work extends explicit reasoning to embeddings: Think-then-Embed (TTE) \cite{cui2025think}, UME-R1 \cite{lan2025ume}, TRACE \cite{hao2026tracetaskadaptivereasoningrepresentation}, and Embed-RL \cite{jiang2026embed} generate an explicit reasoning trace before extracting the embedding, and reasoning-augmented retrieval representations \cite{lan2025llave} further confirm the value of extra reasoning computation. These methods yield consistent accuracy gains, yet producing hundreds of reasoning tokens per sample sharply increases latency and memory cost. In contrast, PLUME retains the benefits of structured reasoning without generating rationale tokens at inference time.

\subsection{Latent Reasoning in Large Language Models}
\begin{figure*}[t]
    \centering
    \includegraphics[width=\textwidth]{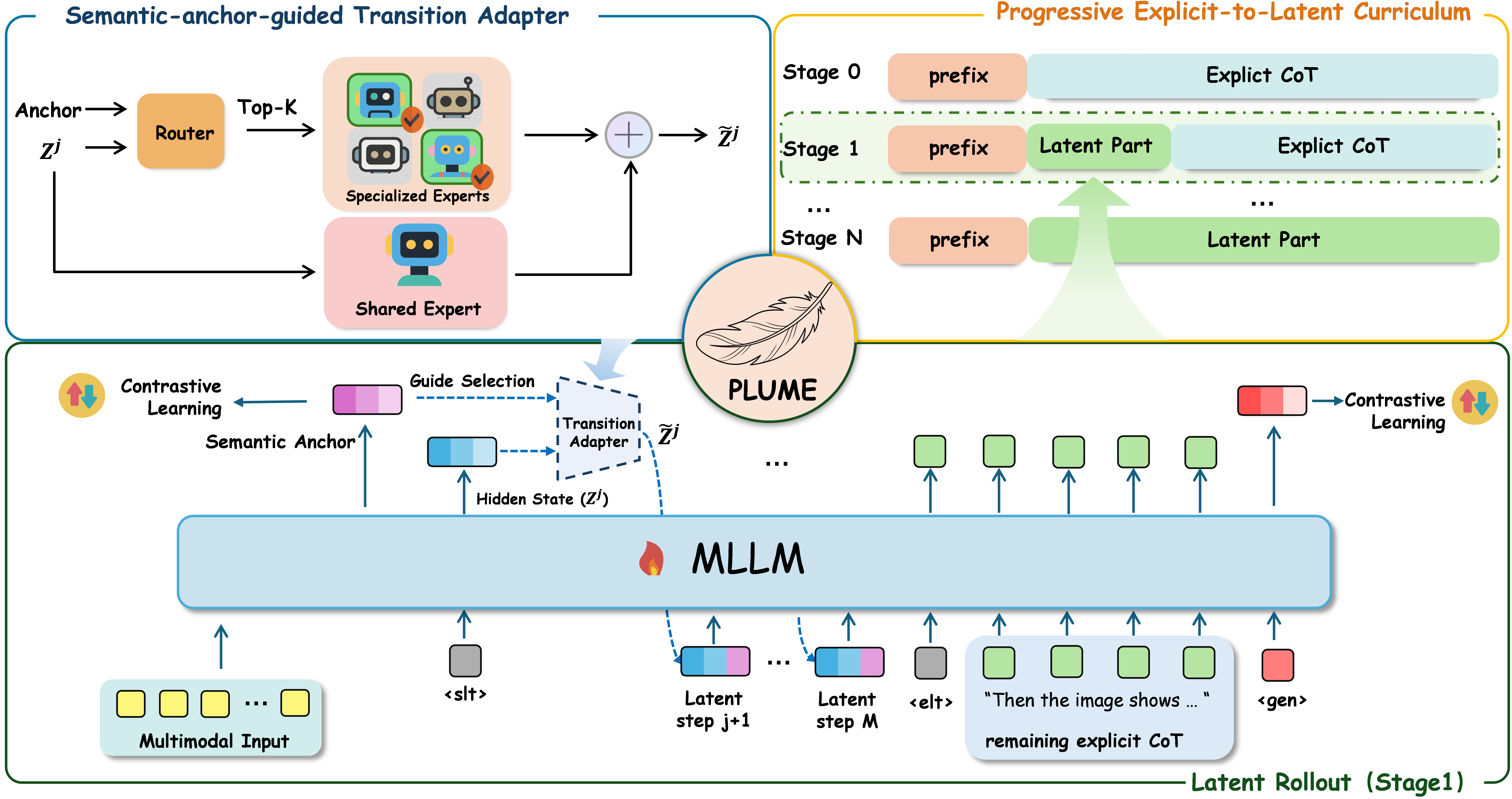}
    \caption{\textbf{Overview of PLUME.} Starting from a multimodal prefix, PLUME replaces explicit CoT decoding with a compact latent rollout inside the backbone. The bottom panel illustrates the latent rollout process, where the model performs several latent transitions before extracting the final retrieval embedding from the hidden state at \texttt{<gen>}. The top-left panel expands the semantic-anchor-guided transition adapter, which routes each latent step through shared and specialized experts, while the top-right panel shows the progressive explicit-to-latent curriculum that gradually rewrites explicit reasoning segments into latent blocks across training stages. The example in the bottom panel corresponds to an intermediate curriculum stage.}
    \label{fig:plume_overview}
\end{figure*}
A parallel line of work studies additional internal computation or latent reasoning beyond explicit CoT \cite{latent_cot_survey}. Pause-token methods \cite{goyal2024think_before_speak} increase internal compute before token prediction, Quiet-STaR \cite{zelikman2024quietstar} trains models to generate useful internal thoughts, and Coconut \cite{coconut} and CODI \cite{codi} move reasoning into continuous hidden space. Fast Quiet-STaR \cite{yang2024fast_quietstar} further compresses thought traces to reduce inference overhead. In retrieval, LaSER \cite{rankovic2025laser} internalizes explicit reasoning into latent space for dense text retrieval. Different from LaSER, which focuses on text-only dense retrieval with a uniform latent reasoning process, PLUME targets universal multimodal embedding, where latent reasoning must operate over heterogeneous inputs and allocate a compact reasoning budget adaptively across diverse query structures.


\section{Method}

\subsection{Overview of PLUME}
\label{sec:method_overview}

PLUME is a progressive latent reasoning framework for universal multimodal embedding. It replaces explicit reasoning tokens with a short latent rollout, adapts each latent transition with a semantic-anchor-guided transition adapter, and transfers explicit reasoning into hidden-space computation through a progressive curriculum. The backbone is fully fine-tuned; the only added components are the lightweight routed adapter and its anchor-conditioned router. Retrieval embeddings are taken directly from normalized hidden states of the backbone, without introducing separate retrieval heads. Figure~\ref{fig:plume_overview} summarizes the framework.

\subsection{Problem Formulation}
\label{sec:problem_formulation}

We consider \emph{universal multimodal embedding} (UME) \cite{jiang2024vlm2vec}, where a single model maps heterogeneous inputs into a shared embedding space for retrieval. Given a query $q$ and its corresponding positive target $t^{+}$, together with a set of negative targets $\mathcal{T}^{-}=\{t_1^{-}, \dots, t_{N_e}^{-}\}$, the goal is to maximize the similarity between $q$ and $t^{+}$ against all negatives. Both queries and targets may be text, images, videos, visual documents, or their combinations.

In practice, we sample a mini-batch of $N$ query--target pairs $\{(q_i,t_i)\}_{i=1}^{N}$, where $(q_i,t_i)$ forms the positive pair and all other targets $\{t_j \mid j \neq i\}$ serve as in-batch negatives for $q_i$. We optimize the model with the InfoNCE objective \cite{oord2018representation}
\begin{equation}
\small
\mathcal{L}_{\mathrm{NCE}}
=
\frac{1}{N}\sum_{i=1}^{N}
-\log
\frac{\exp\!\left(\mathrm{sim}(q_i,t_i)/\tau\right)}
{\sum_{j=1}^{N}\exp\!\left(\mathrm{sim}(q_i,t_j)/\tau\right)},
\end{equation}
where $\mathrm{sim}(\cdot,\cdot)$ denotes cosine similarity between the normalized embeddings produced by the model, and $\tau$ is the temperature hyper-parameter. Unless otherwise specified, we apply this objective bidirectionally in both query-to-target and target-to-query directions. A causal language modeling loss $\mathcal{L}_{\mathrm{CE}}$ is further applied to the decoded suffix of \emph{both} the query and its positive target, providing token-level supervision that grounds the generative pathway, as detailed in Sec.~\ref{sec:embedding_objective}.

The retrieval objective above is standard for UME; PLUME's contribution lies in \emph{how} the embedding is formed---through a compact latent reasoning process within the backbone, described next.

\subsection{Latent Rollout for Universal Multimodal Embedding}
\label{sec:latent_reasoning}

PLUME replaces explicit CoT decoding with a short autoregressive rollout in hidden space---retaining the iterative structure of multi-step reasoning while avoiding the need to materialize intermediate tokens at inference time. Below we describe the four stages of this process: prefix encoding, latent initialization, iterative rollout, and suffix decoding.

\noindent\textbf{Multimodal prefix encoding.}
Given an input $x$, the backbone first processes the multimodal prefix---which may interleave text tokens with image, video, or document features---and then encounters a special \texttt{<slt>} (start-latent-thinking) token that opens a latent block \texttt{<slt>\,<ct>$^{K}$\,<elt>}, where \texttt{<ct>} reserves $K$ autoregressive positions and \texttt{<elt>} (end-latent-thinking) marks the end of the latent block. The \texttt{<ct>} tokens serve only as positional placeholders in the serialized sequence; at each latent step the model receives a continuous hidden state rather than a discrete token embedding.

Let $\mathbf{h}_{1},\ldots,\mathbf{h}_{L}$ denote the hidden states produced by the backbone on the prefix, where $\mathbf{h}_L$ corresponds to \texttt{<slt>}. This prefix pass yields two outputs that persist throughout the latent rollout. The first is the cached key--value states $\mathcal{C}(x)$ \cite{vllm,Vaswani2017AttentionIA}, which allow every subsequent latent step to attend back to the full multimodal context via causal attention. The second is a \emph{semantic anchor} $\mathbf{c}(x)$, extracted from the hidden state at a dedicated \texttt{<anchor>} token in the prefix; it provides a fixed summary of the input's semantic intent for routing (Sec.~\ref{sec:expert_routing}). All visual features---image patches, video frames, and document renderings---are injected during this prefix encoding stage only; the latent rollout proceeds purely in hidden space.

\noindent\textbf{Latent state initialization.}
We initialize the latent state from the hidden state at the \texttt{<slt>} position,
\begin{equation}
    \mathbf{z}^{(0)}=\mathbf{h}_{L},
\end{equation}
since this state already summarizes the multimodal context accumulated before the latent block.

\noindent\textbf{Iterative latent rollout.}
At each latent step $k \in \{1,\ldots,K\}$, PLUME performs two operations. First, the previous latent state $\mathbf{z}^{(k-1)}$ is refined by the routed adapter (Sec.~\ref{sec:expert_routing}), yielding an adapted state $\tilde{\mathbf{z}}^{(k-1)}$. Second, $\tilde{\mathbf{z}}^{(k-1)}$ is fed into the backbone as the input embedding at position $p_{\texttt{<slt>}}+k$, reusing the accumulated KV cache and advancing to the next causal position:
\begin{equation}
    \mathbf{z}^{(k)}=\mathcal{B}_{\theta}\!\left(\tilde{\mathbf{z}}^{(k-1)},\;\mathcal{C}^{(k-1)},\;p_{\texttt{<slt>}}+k\right), \qquad k=1,\ldots,K,
\end{equation}
where $\mathcal{B}_{\theta}$ denotes one forward pass through the full transformer backbone for a single position. The KV cache grows incrementally: $\mathcal{C}^{(0)} = \mathcal{C}(x)$ is initialized from the prefix encoding, and each step appends its own key--value pair, yielding $\mathcal{C}^{(k)} = \mathcal{C}^{(k-1)} \cup \{(\mathbf{k}^{(k)},\mathbf{v}^{(k)})\}$. Consequently, the output $\mathbf{z}^{(k)}$---the last-layer hidden state at position $p_{\texttt{<slt>}}+k$---attends to the full multimodal prefix \emph{and} all preceding latent states $\mathbf{z}^{(1)},\ldots,\mathbf{z}^{(k-1)}$ through standard causal attention over the growing cache. The $K$ outputs $\mathbf{z}^{(1)},\ldots,\mathbf{z}^{(K)}$ constitute the complete latent reasoning trace.

Each latent step occupies the same causal position where an explicit reasoning token would otherwise be decoded. The backbone sees a continuous vector where it would normally see a token embedding, but the attention mask, positional encoding, and KV-cache mechanics are identical to standard autoregressive generation. PLUME therefore preserves the sequential dependency structure of explicit CoT while replacing discrete token generation with a short sequence of continuous hidden-state transitions.

\noindent\textbf{Suffix decoding and embedding extraction.}
After the $K$-step rollout, the latent block is closed by \texttt{<elt>}, and \texttt{<gen>} is placed immediately after \texttt{<elt>}. The final retrieval embedding is extracted from the hidden state at \texttt{<gen>} (Sec.~\ref{sec:embedding_objective}). In practice, PLUME replaces hundreds of explicit reasoning tokens with as few as $K$ latent steps while retaining full KV-cache compatibility.

\subsection{Semantic-Anchor-Guided Transition Adapter}
\label{sec:expert_routing}

A uniform latent transition cannot accommodate the diversity of UME instances, which vary in modality composition, grounding requirements, and reasoning structure. Inspired by mixture-of-experts (MoE) architectures \cite{shazeer2017outrageously,fedus2022switch}, PLUME inserts a lightweight routed adapter between adjacent latent steps, making each transition input-adaptive without modifying the backbone.

Routing is conditioned on the semantic anchor $\mathbf{c}(x)$ (Sec.~\ref{sec:latent_reasoning}), a fixed global signal that stabilizes expert selection against the rapidly evolving latent state.

At latent step $k$, the router concatenates the additively fused state $\mathbf{z}^{(k-1)}+\mathbf{c}(x)$ with a learnable step embedding $\mathbf{e}^{(k)} \in \mathbb{R}^{D}$. The routing weights over $M_e$ specialized experts are:
\begin{equation}
    \boldsymbol{\pi}^{(k)}
    =
    \mathrm{Softmax}\!\left(
    W_r \left[\mathbf{z}^{(k-1)}+\mathbf{c}(x)\,;\,\mathbf{e}^{(k)}\right]+\mathbf{b}_r
    \right),
\end{equation}
where $[\cdot\,;\,\cdot]$ denotes concatenation, $W_r \in \mathbb{R}^{M_e \times 2D}$ and $\mathbf{b}_r \in \mathbb{R}^{M_e}$ are learnable router parameters. Additive fusion injects the anchor signal without altering the dimensionality of the router input, while the step embedding enables the router to distinguish early and late latent steps.

Each expert $E_m$ is a two-layer MLP with an expansion ratio of 2: a linear projection $D \to 2D$ followed by GELU activation \cite{Hendrycks2016GaussianEL}, then a projection back to $D$. A layer normalization is applied to $\mathbf{z}^{(k-1)}$ before it enters any expert, and dropout is applied after the activation for regularization. The adapter contains one shared expert $E_0$ that captures broadly useful transition patterns, plus $M_e$ specialized experts $\{E_m\}_{m=1}^{M_e}$ from which the router selects the top~$K_r$. The adapted latent state combines the shared expert output with a weighted mixture of the selected specialized experts via a residual connection:
\begin{equation}
    \begin{aligned}
        \tilde{\mathbf{z}}^{(k-1)}
        &= \mathbf{z}^{(k-1)} + E_0\!\left(\hat{\mathbf{z}}^{(k-1)}\right) \\
        &\quad +\!\sum_{m \in \mathrm{Top}K_r(\boldsymbol{\pi}^{(k)})}\!
        \pi_m^{(k)}\, E_m\!\left(\hat{\mathbf{z}}^{(k-1)}\right),
    \end{aligned}
\end{equation}
where $\hat{\mathbf{z}}^{(k-1)}=\mathrm{LN}(\mathbf{z}^{(k-1)})$ is the layer-normalized input. The resulting $\tilde{\mathbf{z}}^{(k-1)}$ is passed to the backbone step in Eq.~(3). Thus, routing in PLUME acts only on a lightweight latent transition module, rather than turning the backbone itself into a full MoE model.

To prevent routing collapse, we impose a balance regularizer. Let $\bar{\pi}_m = \frac{1}{NK}\sum_{i=1}^{N}\sum_{k=1}^{K}\pi_{i,m}^{(k)}$ denote the average routing mass for expert~$m$ across the mini-batch and latent steps. The balance loss penalizes deviation from uniform allocation:
\begin{equation}
    \mathcal{L}_{\mathrm{bal}}
    =
    \frac{1}{M_e}\sum_{m=1}^{M_e}
    \left(\bar{\pi}_m - \frac{1}{M_e}\right)^{2},
\end{equation}
which reaches its minimum of zero when all experts receive equal routing mass. This design makes latent reasoning adaptive without sacrificing efficiency: the evolving latent state captures local step-wise computation, while the fixed semantic anchor provides a stable global cue, so that different multimodal inputs follow different latent reasoning paths under the same backbone and rollout budget.

\subsection{Embedding Formation and Training Objectives}
\label{sec:embedding_objective}

The final retrieval embedding is taken from the generative pathway, because it reflects the full latent-to-suffix computation induced by reasoning.

Formally, for an input $x$, we define the final retrieval embedding as
\begin{equation}
    \mathbf{e}_{\mathrm{gen}}(x)=\mathrm{Norm}\!\left(\mathbf{h}_{\texttt{<gen>}}\right),
\end{equation}
where $\mathbf{h}_{\texttt{<gen>}}$ denotes the last-layer hidden state at the \texttt{<gen>} position and $\mathrm{Norm}(\cdot)$ denotes $\ell_2$ normalization. Additionally, an auxiliary anchor embedding $\mathbf{e}_{\mathrm{anc}}(x)=\mathrm{Norm}(\mathbf{h}_{\mathrm{anchor}})$ is derived from the semantic anchor introduced in Sec.~\ref{sec:latent_reasoning}. During training, $\mathbf{e}_{\mathrm{anc}}$ receives its own contrastive supervision ($\mathcal{L}_{\mathrm{NCE}}^{\mathrm{anc}}$), which serves two purposes: it encourages the anchor to encode a semantically meaningful global summary of the input, thereby providing a higher-quality routing signal for the MoE adapter, and it supplies an additional gradient pathway that stabilizes early-stage training when the latent rollout has not yet converged. Crucially, $\mathbf{e}_{\mathrm{anc}}$ is \emph{discarded at inference}: retrieval always relies exclusively on $\mathbf{e}_{\mathrm{gen}}$. Because $\mathbf{e}_{\mathrm{gen}}$ is extracted \emph{after} the full latent rollout and suffix decoding, it cannot be computed without a functioning latent trajectory, which prevents the model from taking a shortcut through the anchor embedding alone.

The full training objective combines suffix generation, retrieval alignment, and routing regularization. The causal language modeling loss $\mathcal{L}_{\mathrm{CE}}$ is computed on the decoded suffix of \emph{both} the query and its positive target, so that the generative pathway receives token-level supervision on both sides of each training pair. We apply the InfoNCE objective \cite{oord2018representation,infonce} in Sec.~\ref{sec:problem_formulation} to both the generative and the anchor embeddings, producing $\mathcal{L}_{\mathrm{NCE}}^{\mathrm{gen}}$ and $\mathcal{L}_{\mathrm{NCE}}^{\mathrm{anc}}$, respectively. The overall objective is
\begin{equation}
    \mathcal{L}
    =
    \mathcal{L}_{\mathrm{CE}}
    + \lambda_{\mathrm{gen}} \mathcal{L}_{\mathrm{NCE}}^{\mathrm{gen}}
    + \lambda_{\mathrm{anc}} \mathcal{L}_{\mathrm{NCE}}^{\mathrm{anc}}
    + \lambda_{\mathrm{bal}} \mathcal{L}_{\mathrm{bal}},
\end{equation}
where $\lambda_{\mathrm{gen}}$, $\lambda_{\mathrm{anc}}$, and $\lambda_{\mathrm{bal}}$ are balancing coefficients.

The generative pathway thus receives primary embedding supervision, while the anchor pathway provides auxiliary training signal that improves routing quality and training stability; the anchor embedding itself is discarded at inference, and the retrieval embedding depends exclusively on the latent rollout.

\subsection{Progressive Explicit-to-Latent Curriculum}
\label{sec:progressive_curriculum}

A direct transition from explicit CoT supervision to latent-only execution is unstable, because semantic grounding does not transfer reliably into hidden-space rollout. Without an explicit scaffold, latent states may collapse into degenerate shortcuts instead of preserving the multi-step structure learned from verbalized reasoning. PLUME therefore adopts a progressive explicit-to-latent curriculum \cite{ICML-2009-BengioLCW}, which uses explicit CoT only as a transient training scaffold.

\begin{table*}[t]
\centering
\small
\setlength{\tabcolsep}{3.6pt}
\renewcommand{\arraystretch}{1.10}
\caption{Main comparison on MMEB-v2. The first seven models use Qwen2-VL-2B and the final two use Qwen2-VL-7B. Within each backbone size, \textbf{bold} and \underline{underlining} mark the best and second-best scores, respectively.}
\label{tab:main_results}
\resizebox{\textwidth}{!}{
\begin{tabular}{llcccccccccccccccc}
\toprule
\multirow{2}{*}{Model}
& \multirow{2}{*}{\textcolor{gray}{Venue}}
& \multicolumn{5}{c}{Image}
& \multicolumn{5}{c}{Video}
& \multicolumn{5}{c}{VisDoc}
& \multirow{2}{*}{All} \\
\cmidrule(lr){3-7} \cmidrule(lr){8-12} \cmidrule(lr){13-17}
&
& CLS & QA & RET & GD & Overall
& CLS & QA & RET & MRET & Overall
& VDRv1 & VDRv2 & VR & OOD & Overall
& \\
\midrule
\# of Datasets
& \textcolor{gray}{--}
& 10 & 10 & 12 & 4 & 36
& 5 & 5 & 5 & 3 & 18
& 10 & 4 & 6 & 4 & 24
& 78 \\
\midrule

\rowcolor{gray!18}
\multicolumn{18}{c}{\textit{Early UME Baselines}} \\
LamRA
& \textcolor{gray}{CVPR'25}
& 59.2 & 26.5 & \textbf{70.0} & 62.7 & 54.1 & 39.3 & 42.6 & 24.3 & 34.6 & 35.2 & 22.0 & 11.5 & 37.4 & 21.0 & 23.9 & 40.4 \\
VLM2Vec
& \textcolor{gray}{ICLR'25}
& 58.7 & 49.3 & 65.0 & 72.9 & 59.7 & 33.4 & 30.5 & 20.6 & 33.0 & 29.0 & 49.8 & 13.5 & 51.8 & 33.5 & 41.6 & 47.0 \\
GME
& \textcolor{gray}{CVPR'25}
& 54.4 & 29.9 & 66.9 & 55.5 & 51.9 & 34.9 & 42.0 & 25.6 & 32.4 & 33.9 & \textbf{86.1} & \textbf{54.0} & \textbf{82.5} & \underline{43.1} & \textbf{72.7} & 54.1 \\
VLM2Vec-V2
& \textcolor{gray}{TMLR'26}
& 62.0 & 56.3 & \underline{69.5} & 77.3 & 64.9 & 39.3 & 34.3 & 28.8 & 38.5 & 34.9 & \underline{75.5} & 44.9 & \underline{79.4} & 39.4 & 65.4 & 58.0 \\
DUME (2B, w/o CoT)
& \textcolor{gray}{ICLR'26}
& 59.3 & 55.0 & 66.3 & \underline{78.0} & 62.5 & 37.7 & 46.6 & 17.1 & 30.0 & 33.2 & 67.6 & 43.3 & 47.1 & 33.8 & 52.8 & 52.7 \\

\midrule
\rowcolor{gray!18}
\multicolumn{18}{c}{\textit{Reasoning UME}} \\
UME-R1 (2B)
& \textcolor{gray}{ICLR'26}
& \underline{64.8} & \textbf{62.8} & 67.6 & 77.2 & \textbf{66.6} & \underline{44.3} & \underline{51.2} & \underline{32.9} & \underline{39.7} & \underline{42.2} & 72.4 & 46.2 & 79.2 & 37.2 & 63.9 & \underline{60.1} \\
\rowcolor{yellow!15}
PLUME (2B)
& \textcolor{gray}{Ours}
& \textbf{66.5} & \underline{59.2} & 67.6 & \textbf{79.7} & \underline{66.3}
& \textbf{45.0} & \textbf{52.3} & \textbf{33.5} & \textbf{46.7} & \textbf{44.1}
& 72.1 & \underline{49.8} & 78.1 & \textbf{57.4} & \underline{67.5} & \textbf{61.6} \\
\midrule
UME-R1 (7B)
& \textcolor{gray}{ICLR'26}
& \underline{69.5} & \textbf{67.5} & \underline{72.3} & \underline{81.9} & \underline{71.3}
& \underline{49.5} & \underline{56.4} & \underline{38.1} & \underline{44.9} & \underline{47.5}
& \underline{75.6} & \underline{49.4} & \underline{82.4} & \underline{40.4} & \underline{67.1} & \underline{64.5} \\
\rowcolor{yellow!15}
PLUME (7B)
& \textcolor{gray}{Ours}
& \textbf{71.9} & \underline{64.6} & \textbf{73.0} & \textbf{85.1} & \textbf{71.7}
& \textbf{51.2} & \textbf{58.5} & \textbf{39.7} & \textbf{52.9} & \textbf{50.3}
& \textbf{77.1} & \textbf{54.8} & \textbf{83.1} & \textbf{62.4} & \textbf{72.4} & \textbf{67.1} \\

\bottomrule
\end{tabular}
}
\end{table*}

For each training instance, we split the explicit rationale into sentence-level segments and gradually replace them, from left to right, with a latent block of \texttt{<ct>} positions. At early stages, most reasoning steps remain explicit and are teacher-forced, which preserves semantic grounding. As training proceeds, a larger prefix of the rationale is absorbed into the latent block, while the remaining unreplaced steps are kept as supervised suffix tokens after \texttt{<elt>}. The model thus learns to continue explicit reasoning from partially latent prefixes before internalizing the full reasoning process.

The latent block itself receives no token-level supervision. Supervision is applied only to the remaining explicit rationale and the downstream answer. In the final stage, both the explicit rationale and the answer span are removed, so that the latent rollout connects directly to \texttt{<gen>}, matching the inference-time execution pattern.

We allocate more training to the final fully latent stage, while earlier stages serve mainly to stabilize the transfer from verbalized to latent reasoning. This curriculum provides a structured bridge from explicit CoT to compact latent rollout, allowing PLUME to retain the benefits of structured intermediate reasoning without preserving explicit CoT at inference time.

\section{Experiments}

\subsection{Experimental Setup}
\label{sec:experimental_setup}

\noindent\textbf{Evaluation Metrics.}
We evaluate PLUME on MMEB-v2 \cite{meng2025vlm2vec}, a comprehensive benchmark for universal multimodal embedding. MMEB-v2 extends MMEB-V1 \cite{jiang2024vlm2vec} by introducing video and visual-document retrieval scenarios, resulting in a benchmark with 9 meta-tasks and 78 test tasks in total. The benchmark covers a broad spectrum of vision-language retrieval settings, including image-level retrieval, video understanding and retrieval, visual-document retrieval, and reasoning-intensive multimodal matching. Following prior work, we report Hit@1 for image and video tasks, and NDCG@5 \cite{jarvelin2002cumulated} for visual-document retrieval tasks. 

\noindent\textbf{Baselines.}
We compare PLUME against two groups of representative baselines. The first group consists of early UME methods that form embeddings without explicit reasoning, including LamRA \cite{liu2025lamra}, VLM2Vec \cite{jiang2024vlm2vec}, GME \cite{zhang2024gme}, VLM2Vec-V2 \cite{meng2025vlm2vec}, and DUME (UME-R1 without CoT) \cite{lan2025ume}. The second group consists of reasoning-enhanced UME methods, represented by UME-R1 \cite{lan2025ume}, which generates explicit CoT rationales before embedding extraction. Table~\ref{tab:main_results} compares these methods on Qwen2-VL-2B \cite{wang2024qwen2}. To test whether the gains extend to a larger backbone, we also compare PLUME and UME-R1 on Qwen2-VL-7B. We note that concurrent work TTE \cite{cui2025think} employs a separate Qwen2.5-VL-72B \cite{bai2025qwen25vl} model as a dedicated reasoning module to produce CoT rationales before the 2B encoder extracts the final embedding; because this introduces additional large-scale model capacity that is unavailable to the other methods, we do not include it in the main comparison. We similarly exclude methods built on different backbone families or training corpora.

\noindent\textbf{Implementation Details.}
Our primary model uses the Qwen2-VL-2B backbone \cite{wang2024qwen2}. For training data, we use the same supervised fine-tuning corpus as UME-R1 \cite{lan2025ume}, since it provides explicit multimodal reasoning traces that are well suited for progressive explicit-to-latent transfer. The InfoNCE temperature is set to 0.02, the global batch size is 1024, and the learning rate is $5\times10^{-5}$. The latent rollout length is set to $K\!=\!8$. Training lasts for 5 epochs in total. The curriculum begins with an initial fully explicit-CoT warm-up stage (Stage 0) trained for 3 epochs, followed by four progressive latent curriculum stages (Stages 1--4). Among them, the intermediate transition stages (Stages 1--3) are completed within 1 epoch in total, and the final fully latent stage (Stage 4) is trained for 1 additional epoch. The 7B comparison uses the same training data, evaluation protocol, and retrieval setting as the 2B comparison.

\subsection{Main Comparison on MMEB-v2}
\label{sec:main_results}

Table~\ref{tab:main_results} compares single-pass and reasoning-enhanced methods at 2B and reports the size-matched comparison between PLUME and UME-R1 at 7B.

With the same 2B backbone and training data, PLUME surpasses UME-R1 \cite{lan2025ume} by 1.5 points overall (61.6 vs.\ 60.1) while requiring only 8 latent steps instead of hundreds of generated reasoning tokens. Compared with early baselines, PLUME surpasses the strongest single-pass method VLM2Vec-V2 \cite{meng2025vlm2vec} by 3.6 points overall, with particularly large gains on Video (+9.2) where multi-step temporal reasoning is most beneficial.

\begin{figure}[t]
    \centering
    \includegraphics[width=\linewidth]{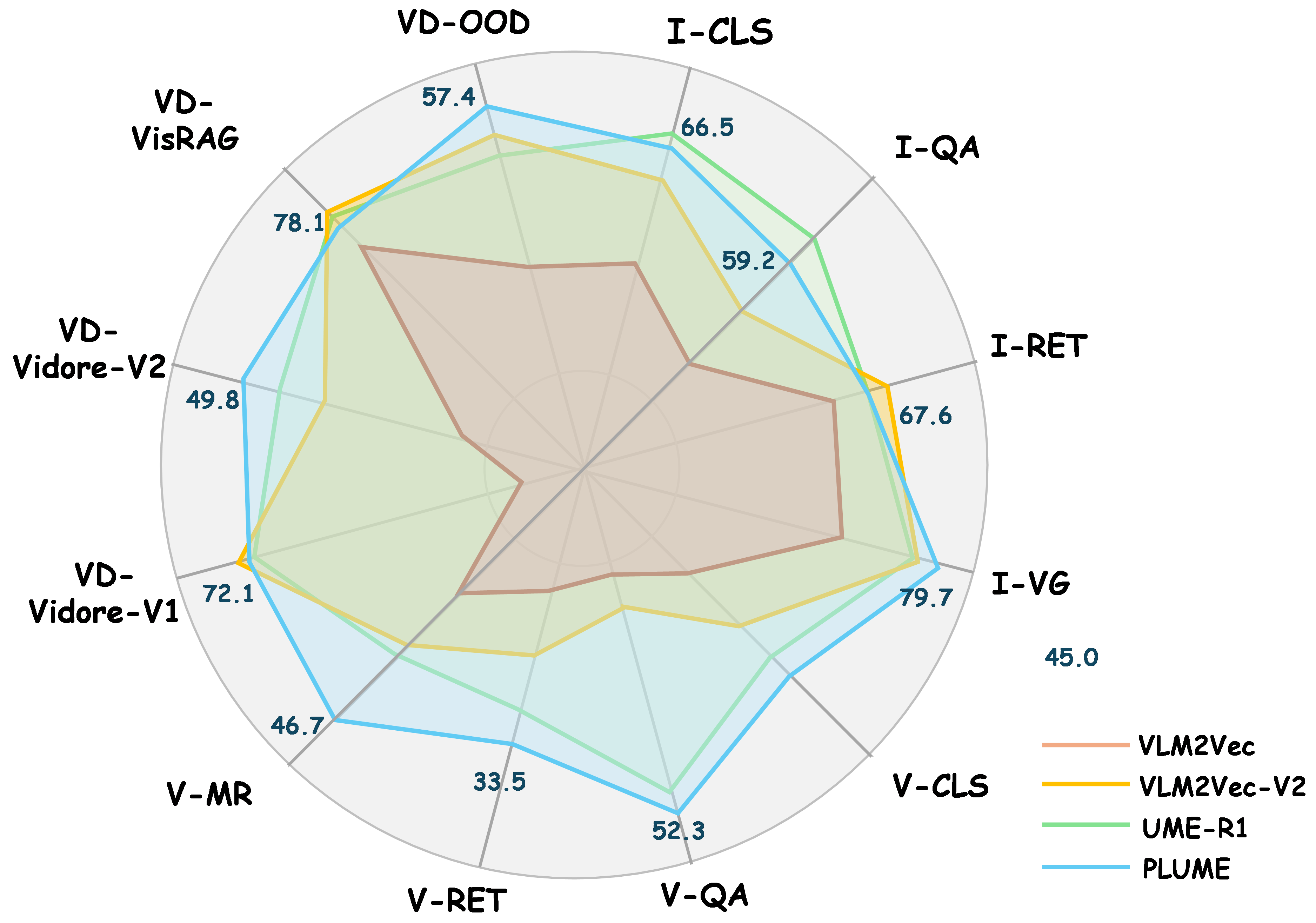}
    \caption{\textbf{Per-task performance of the 2B models on MMEB-v2.}}
    \label{fig:rader}
\end{figure}

Across modality groups, PLUME (2B) achieves 66.3 on Image (vs.\ UME-R1's 66.6), 44.1 on Video (vs.\ 42.2), and 67.5 on VisDoc (vs.\ 63.9). While PLUME trails UME-R1 slightly on Image ($-0.3$), it delivers clear gains on Video (+1.9) and VisDoc (+3.6). The Video advantage is consistent with our motivation that temporal dynamics and cross-frame relationships are difficult to linearize into discrete tokens, and that maintaining continuous state across reasoning steps preserves richer temporal semantics. This advantage is most visible on Video Multi-modal Retrieval (PLUME 46.7 vs.\ UME-R1 39.7, +7.0), where multi-step cross-modal alignment benefits from uninterrupted continuous reasoning. PLUME also sets the best score on Image Grounding (79.7) and VisDoc OOD (57.4), both of which involve compositional spatial reasoning or out-of-distribution generalization that benefit from continuous-state computation. Figure~\ref{fig:rader} visualizes these per-task comparisons, confirming that PLUME consistently outperforms UME-R1 and single-pass baselines across most sub-tasks.

With Qwen2-VL-7B, PLUME scores 67.1 overall compared with 64.5 for UME-R1 (+2.6 points). PLUME gains 0.4 points on Image (71.7 vs.\ 71.3), 2.8 on Video (50.3 vs.\ 47.5), and 5.3 on VisDoc (72.4 vs.\ 67.1). Image QA is the only 7B sub-task on which UME-R1 scores higher (67.5 vs.\ 64.6).

\subsection{Efficiency and Accuracy-Efficiency Tradeoff}
\label{sec:efficiency}


Table~\ref{tab:efficiency} provides a quantitative breakdown of inference costs. All measurements use 2B models on a single NVIDIA H20 GPU. For each method, we randomly sample 500 inputs per modality as one evaluation set, preceded by 20 warm-up iterations, and repeat with 5 independently drawn evaluation sets. The table reports the mean across the 5 runs; $\pm$ denotes the standard deviation.

\begin{table}[t]
\centering
\small
\caption{Inference efficiency of the 2B models on a single H20 GPU.}
\label{tab:efficiency}
\resizebox{\linewidth}{!}{

\begin{tabular}{lccc}
\toprule
Metric & PLUME & UME-R1 & VLM2Vec-V2 \\
\midrule
Reasoning tokens/steps & 8 & 403 & 0 \\
Latency (ms/sample) & 298$\pm$12 & 9023$\pm$187 & 156$\pm$8 \\
Throughput (samples/s) & 3.3$\pm$0.1 & 0.11$\pm$0.01 & 6.4$\pm$0.3 \\
\midrule
Speedup vs.\ UME-R1 & 30.3$\times$ & 1.0$\times$ & -- \\
Overhead vs.\ VLM2Vec-V2 & 1.9$\times$ & -- & 1.0$\times$ \\
\bottomrule
\end{tabular}
}
\end{table}

PLUME compresses reasoning from an average of 403 generated tokens (UME-R1 \cite{lan2025ume}) to 8 latent steps, reducing per-sample latency from 9023\,ms to 298\,ms, a 30.3$\times$ speedup. Compared with the single-pass baseline VLM2Vec-V2 \cite{meng2025vlm2vec} (156\,ms), PLUME adds less than 150\,ms of overhead while improving the overall score by 3.6 points (61.6 vs.\ 58.0), confirming that a small latent computation budget delivers substantial reasoning gains at modest cost. 

Figure~\ref{fig:pareto} visualizes the accuracy--efficiency tradeoff. PLUME occupies a favorable position on the Pareto frontier: at $K\!=\!6$ it already surpasses UME-R1's accuracy while running over 30x faster, and at $K\!=\!4$ it still outperforms all single-pass baselines with throughput comparable to VLM2Vec-V2. This confirms that latent reasoning effectively reconciles retrieval quality and inference efficiency.

\subsection{Ablation Studies}
\label{sec:ablation}

We conduct ablation studies with the 2B backbone to verify the contribution of each core component. Unless otherwise stated, all ablations use the same training recipe as the full model. 

\begin{table}[t]
\centering
\small
\caption{Ablation on the core components of PLUME.}
\label{tab:ablation_core}
\setlength{\tabcolsep}{4pt}
\begin{tabular}{lcccc}
\toprule
Configuration & Image & Video & VisDoc & All \\
\midrule
Full PLUME & 66.3 & 44.1 & 67.5 & 61.6 \\
w/o Latent Transition & 63.6 & 41.0 & 64.8 & 58.8 \\
w/o MoE (single MLP) & 64.2 & 41.8 & 64.4 & 59.2 \\
w/o Semantic Anchor & 65.4 & 42.3 & 66.1 & 60.1 \\
w/o Curriculum & 60.2 & 36.5 & 60.2 & 54.8 \\
\bottomrule
\end{tabular}
\end{table}

\noindent\textbf{Component ablation.}
Table~\ref{tab:ablation_core} shows that every component contributes meaningfully.
Removing the progressive curriculum causes the largest overall drop ($-6.8$), with Video suffering the most ($-7.6$). In this ablation, the model first completes Stage~0 training with full explicit CoT and then directly trains with the Stage~4 setting (fully latent, no explicit tokens), bypassing the gradual transition of Stages~1--3. The large degradation confirms that an abrupt switch from explicit to latent reasoning leads to training instability, and the progressive schedule is necessary for stable knowledge transfer.

Removing the latent transition entirely (reading the embedding from the last prefix token) reduces accuracy by 2.8 overall, with consistent losses across all modalities, validating that iterative hidden-space computation provides genuine reasoning benefit beyond additional parameters.
Replacing the MoE adapter with a single shared MLP costs 2.4 points, and the degradation is most pronounced on VisDoc ($-3.1$), where document understanding benefits from specialized expert pathways.
Removing the semantic anchor from the router hurts Video most ($-1.8$), indicating that global input context is important for routing temporal reasoning.

\begin{table}[t]
\centering
\small
\caption{Effect of the latent steps $K$ on accuracy and latency.}
\label{tab:ablation_k}
\setlength{\tabcolsep}{3.5pt}
\begin{tabular}{cccccc}
\toprule
$K$ & Image & Video & VisDoc & All & Latency (ms) \\
\midrule
4  & 64.3 & 43.3 & 65.7 & 59.9 & 232 \\
6  & 65.9 & 43.6 & 66.7 & 61.1 & 268 \\
8  & 66.3 & 44.1 & 67.5 & 61.6 & 300 \\
\bottomrule
\end{tabular}
\end{table}

\noindent\textbf{Latent steps $K$.}
Table~\ref{tab:ablation_k} shows that accuracy improves steadily from $K\!=\!4$ to $K\!=\!8$, gaining 1.7 points overall. The gain from 6 to 8 (+0.5) is smaller than from 4 to 6 (+1.2), exhibiting diminishing returns. Latency grows roughly linearly (232\,ms $\to$ 300\,ms), so $K\!=\!8$ offers the best absolute accuracy while $K\!=\!6$ provides a favorable accuracy--speed balance (see also Figure~\ref{fig:pareto}).

\begin{table}[t]
\centering
\small
\caption{Ablation on the transition adapter design.}
\label{tab:ablation_moe}
\setlength{\tabcolsep}{4pt}
\resizebox{\linewidth}{!}{
\begin{tabular}{lcccc}
\toprule
Configuration & Image & Video & VisDoc & All \\
\midrule
Default ($M_e\!=\!4$, $K_r\!=\!2$, shared) & 66.3 & 44.1 & 67.5 & 61.6 \\
w/o Shared Expert & 65.4 & 42.5 & 66.1 & 60.3 \\
Top-1 expert (instead of Top-2) & 65.8 & 43.0 & 66.8 & 60.8 \\
Router: w/o $c(x)$   & 65.4 & 42.3 & 66.1 & 60.1 \\
Router: w/o $e^{(k)}$  & 65.7 & 41.9 & 66.2 & 60.4 \\
\bottomrule
\end{tabular}
}
\end{table}

\noindent\textbf{Transition adapter design.}
Table~\ref{tab:ablation_moe} validates individual design choices in the routed adapter.
Removing the shared expert drops overall accuracy by 1.3, confirming that a modality-agnostic baseline pathway complements the specialized experts.
Top-2 routing outperforms top-1 by 0.8, indicating that combining two experts captures richer transition patterns.
Among routing inputs, removing the semantic anchor $c(x)$ hurts more ($-1.5$) than removing the step embedding $e^{(k)}$ ($-1.2$), yet both contribute: $c(x)$ provides global input context for expert selection, while $e^{(k)}$ encodes positional progression within the rollout.


\subsection{Diagnostic Analysis}
\label{sec:analysis}

\begin{figure}[t]
    \centering
    \includegraphics[width=\linewidth]{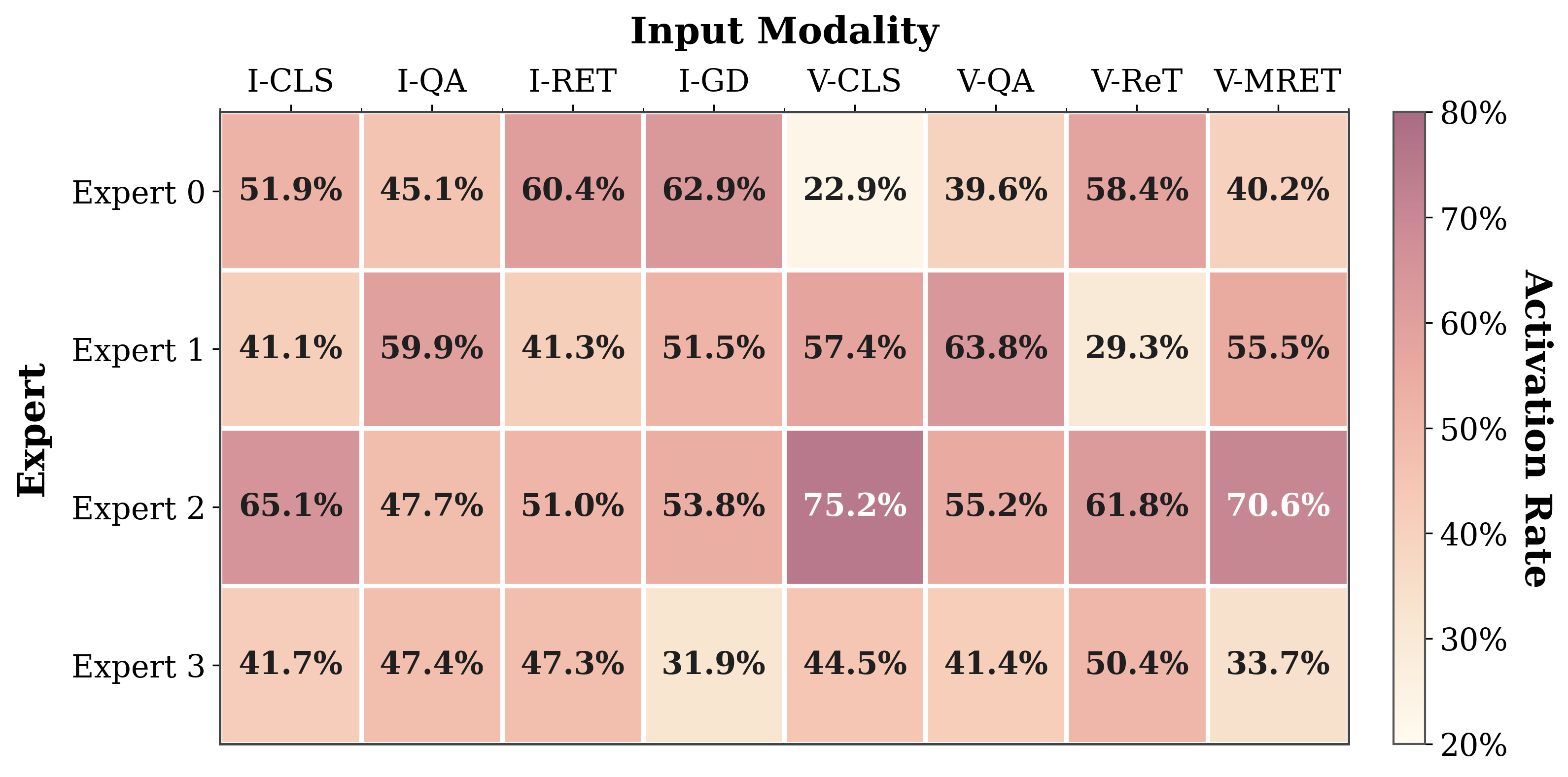}
    \caption{Activation preferences of specialized experts across image and video retrieval sub-tasks.}
    \label{fig:expert_task}
\end{figure}

We visualize the routing behavior of the four specialized experts to examine whether the routed adapter learns meaningful specialization. The shared expert is always active for every input and is therefore omitted from the heatmap; the figure shows only the top-$K_r$ selected specialized experts.

\noindent\textbf{Task-level routing.}
Figure~\ref{fig:expert_task} provides a fine-grained view across image and video sub-tasks.
Expert~2 shows the highest activation on video classification (V-CLS: 75.2\%) and video multi-modal retrieval (V-MRET: 70.6\%), and remains elevated on image classification (I-CLS: 65.1\%) and video retrieval (V-ReT: 61.8\%).
Expert~1 is preferentially activated on question-answering tasks: V-QA (63.8\%) and I-QA (59.9\%), consistent with an affinity for knowledge-intensive reasoning.
Expert~0 peaks on image grounding (I-GD: 62.9\%) and image retrieval (I-RET: 60.4\%), while being almost never selected for video classification (V-CLS: 22.9\%).
Expert~3 remains a low-activation generalist without strong task-level peaks.
These activation patterns emerge purely from the routing objective without any explicit task labels, confirming that the semantic-anchor-guided router adapts latent computation to the structural demands of each input.

\begin{figure}[t]
    \centering
    \includegraphics[width=\linewidth]{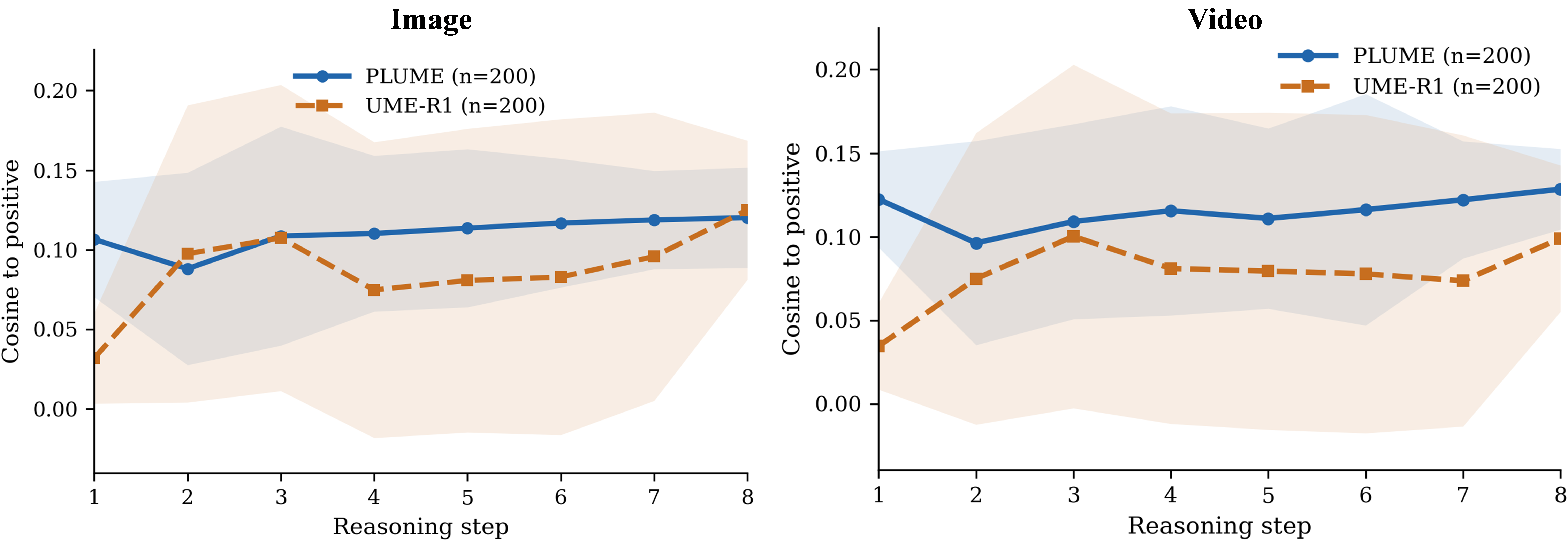}
    \caption{Average cosine similarity between intermediate states and the positive target over 200 samples, reported separately on image and video retrieval. PLUME shows a smoother trajectory with consistently smaller variance than UME-R1 across reasoning steps.}
    \label{fig:pca_trajectory}
\end{figure}

\noindent\textbf{Latent trajectory visualization.}
Figure~\ref{fig:pca_trajectory} compares the average cosine similarity between intermediate states and the positive target over 200 samples, reported for image and video retrieval. We use this metric as a diagnostic signal for trajectory stability rather than as a requirement that every intermediate step must monotonically approach the positive target. Across both subsets, PLUME exhibits a smoother trajectory with consistently smaller variance than UME-R1, especially after the early reasoning steps. On image retrieval, the two methods are close at the beginning, but UME-R1 shows a larger mid-trajectory drop and substantially broader dispersion, whereas PLUME remains more stable throughout the rollout. On video retrieval, the advantage is clearer: PLUME maintains stronger alignment with the positive target across most steps, while UME-R1 stays lower and fluctuates more. These trends suggest that latent reasoning provides a more consistent intermediate computation path for retrieval, whereas explicit CoT produces more variable hidden-state trajectories under discrete token generation.

\noindent\textbf{Limitations.}
Despite surpassing UME-R1 overall, PLUME shows a notable gap on the Image QA subset. This weakness is not uniform across all QA tasks: the gap is small on ScienceQA and WebQA, but much larger on text-rich or knowledge-intensive benchmarks such as ChartQA, InfographicsVQA and OK-VQA. We hypothesize that these tasks rely more heavily on preserving fine grained textual detail and explicit intermediate semantic organization, whereas PLUME compresses reasoning into a short latent rollout optimized for retrieval-oriented representation formation. Besides, while the routed adapter develops differentiated activation patterns consistent with our design hypothesis, formal interpretability guarantees for continuous latent trajectories remain an open problem.

\section{Conclusion}
We introduced PLUME, a latent reasoning framework for universal multimodal embedding that replaces explicit chain-of-thought generation with a short hidden-space rollout. By combining latent multi-step reasoning, anchor-guided routed adaptation, and a progressive explicit-to-latent curriculum, PLUME achieves better transfer of reasoning into compact embeddings. On the 78-task MMEB-v2 benchmark, PLUME improves on size-matched UME-R1 by 1.5 points at 2B and 2.6 points at 7B. With the 2B backbone, eight latent steps replace an average of 403 generated tokens, yielding a $30.3\times$ inference speedup.

{
    \small
    \bibliographystyle{ieeenat_fullname}
    \bibliography{main}
}


\clearpage

\appendix

\section{Curriculum Ablation Details}
\label{app:curriculum_ablation}

We provide additional ablations on the curriculum design used in PLUME. Specifically, we vary the number of curriculum stages while keeping the backbone, training data, total number of training epochs, and all other optimization settings fixed. Unless otherwise specified, all variants use the same final latent rollout length and differ only in how progressively the explicit reasoning segments are rewritten into latent computation.

Table~\ref{tab:curriculum_stage_ablation} compares curriculum schedules with 2, 4, 6, and 8 stages. Overall, we observe that curriculum granularity has a non-monotonic effect on performance. Using only 2 stages yields the weakest result, suggesting that an overly abrupt transition from explicit reasoning supervision to latent computation is suboptimal. With too few stages, the model must absorb a large distribution shift at each transition, which makes it harder to preserve stable representation quality during the handoff from token-level reasoning traces to latent reasoning steps.

Increasing the number of stages to 4 substantially improves performance across all three domains, indicating that a moderately progressive curriculum better supports optimization. A finer curriculum allows the model to adapt more smoothly as explicit reasoning is gradually replaced by latent rollout, reducing optimization difficulty and stabilizing embedding formation. This effect is particularly visible in the visual document setting, where the gap between 2 and 4 stages is the largest.

Further increasing the number of stages beyond 4 does not bring additional gains. Although 6 stages remains competitive, its overall result is slightly below the 4-stage setting, and 8 stages degrades more clearly. Since the total training budget is fixed for all variants, increasing the number of stages necessarily reduces the effective training time allocated to each individual stage. As a result, later stages may be under-optimized before the curriculum advances again, preventing the model from fully adapting to each intermediate supervision regime. In addition, an overly fine-grained curriculum weakens the distinction between adjacent stages, which may reduce the practical benefit of each transition while introducing extra scheduling complexity.

Taken together, these results suggest that the curriculum should be progressive enough to avoid a sharp explicit-to-latent shift, but not so fragmented that each stage becomes too short to be fully exploited. We therefore adopt 4 stages as the default setting, as it provides the best overall trade-off between retrieval accuracy and curriculum efficiency in our experiments.

\begin{table}[t]
    \centering
    \small
    \setlength{\tabcolsep}{6pt}
    \caption{Ablation on the number of curriculum stages. All models are trained with the same data, backbone, and total training recipe unless otherwise specified. The best result is in bold and the second best is underlined.}
    \label{tab:curriculum_stage_ablation}
    \begin{tabular}{lcccc}
        \toprule
        \textbf{\# Stages} & \textbf{Image} & \textbf{Video} & \textbf{VisDoc} & \textbf{Avg.} \\
        \midrule
        2 & 64.0 & 42.1 & 64.3 & 59.0 \\
        4 & 66.3 & \textbf{44.1} & \textbf{67.5} & \textbf{61.6} \\
        6 & \textbf{66.4} & \underline{43.7} & \underline{67.3} & \underline{61.4} \\
        8 & 65.6 & 43.6 & 66.7 & 60.9 \\
        \bottomrule
    \end{tabular}
\end{table}

\textbf{Implementation details.}
For all curriculum variants, we preserve the same overall training setup and only modify the number of transition stages. The rewritten portion of the reasoning sequence increases progressively across stages, while the latent rollout budget is scaled accordingly to match the increasing degree of latent computation. This design isolates the effect of curriculum granularity without confounding it with changes in model capacity or total optimization budget.

\section{Training Time and Computational Cost Analysis}
\label{app:training_cost}

We report the training cost of PLUME-2B under different latent rollout lengths. Since all variants share the same backbone and training recipe, the main difference in computational cost comes from the number of latent reasoning steps.
In our experiments, training PLUME with 4 latent steps requires approximately 2562 H20 GPU hours. The cost increases to about 2838 H20 GPU hours for 6 latent steps and 3119 H20 GPU hours for 8 latent steps. This increase is expected, as longer latent rollouts introduce additional computation during training.
These results show that the training cost grows steadily with the latent rollout length, without introducing unexpected overhead beyond the added latent computation itself.

\section{Failure Case Analysis}
\label{app:failure_cases}

We provide representative failure cases to better understand the remaining limitations of PLUME. In general, PLUME performs strongly on retrieval tasks that benefit from compact multi-step reasoning, but it can still struggle in cases requiring fine-grained textual preservation, dense document understanding, or externally grounded factual knowledge.

\begin{figure*}[!t]
    \centering
    \includegraphics[width=0.96\textwidth]{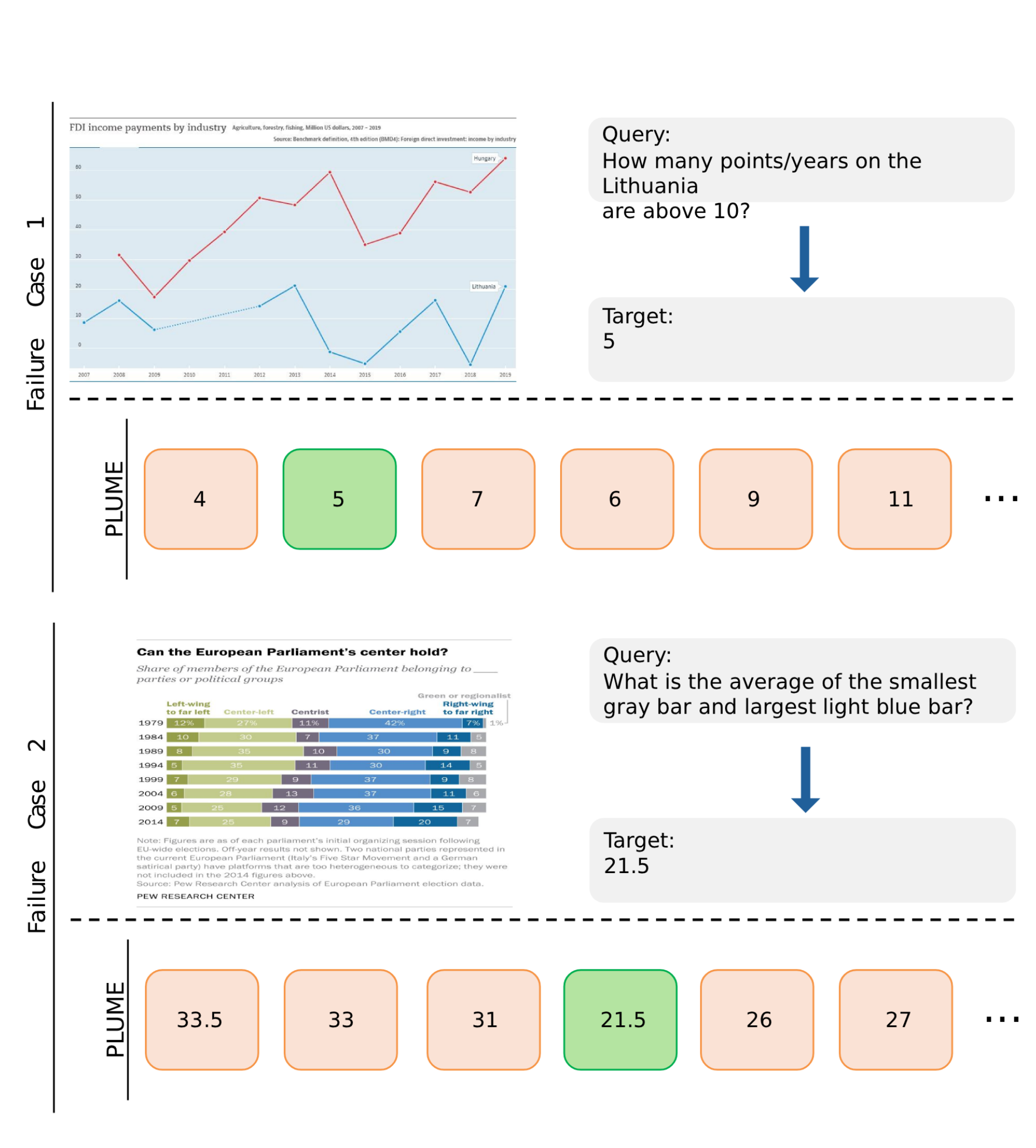}
    \caption{\textbf{Failure cases on ChartQA.}
    Each example shows the input chart and query at the top, and the top retrieved predictions from PLUME at the bottom. The green box marks the ground-truth target, while the preceding orange boxes denote higher-ranked distractors. In Case 1, PLUME fails on a count-over-threshold query, ranking \emph{4} above the correct answer \emph{5}. In Case 2, PLUME struggles with multi-step numerical aggregation, ranking several nearby values ahead of the correct average \emph{21.5}. These examples suggest that although PLUME usually preserves the coarse semantic intent of the query, compressing reasoning into a short latent rollout can still weaken the preservation of fine-grained numerical relations and localized textual details required by chart-intensive question answering.}
    \label{fig:failure_chartqa}
\end{figure*}

Figure~\ref{fig:failure_chartqa} presents representative failure cases from ChartQA.
We focus on this benchmark because it requires fine-grained numerical reading, localized text grounding, and compositional reasoning over structured visual content.
In both examples, the correct answer remains semantically close to the retrieved candidates, but PLUME fails to rank it first.
This pattern suggests that the main limitation is not a complete loss of task intent, but rather insufficient preservation of small numerical distinctions and multi-step relational structure after explicit reasoning is compressed into a compact latent rollout.

\begin{figure*}[!t]
    \centering
    \includegraphics[width=0.96\textwidth]{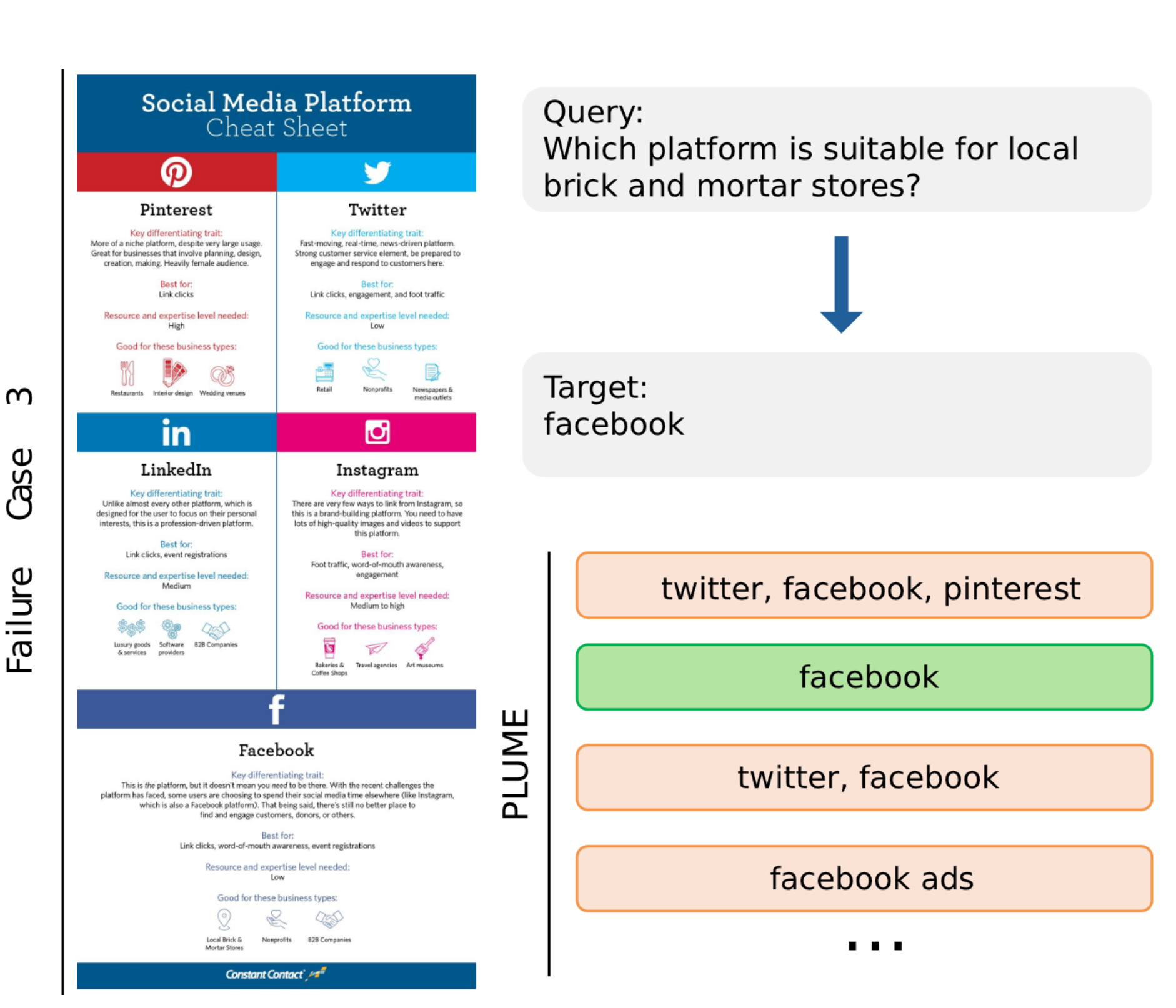}
    \caption{\textbf{Failure case on InfographicsVQA.}
    The figure shows an infographic-style input, the query, the ground-truth target, and the top retrieved predictions from PLUME. The green box marks the correct target, while the orange boxes denote higher-ranked or nearby distractors. In this example, the target answer is \emph{facebook}, but PLUME ranks a broader candidate, \emph{twitter, facebook, pinterest}, above the exact target. It also retrieves semantically related but overly coarse or mismatched variants such as \emph{twitter, facebook} and \emph{facebook ads}. This case suggests that on text-dense infographic inputs, PLUME can capture the relevant semantic region but may fail to preserve the precise answer granularity needed for exact retrieval.}
    \label{fig:failure_infovqa}
\end{figure*}

Figure~\ref{fig:failure_infovqa} presents a representative failure case from InfographicsVQA.
Compared with natural image question answering, infographic style inputs require the model to aggregate fine-grained textual cues distributed across multiple regions under a complex visual layout.
In this example, PLUME correctly identifies the relevant semantic neighborhood, but ranks broader or compositionally related answers ahead of the exact target.
This suggests that the main challenge lies not in coarse semantic localization, but in maintaining precise answer granularity when dense textual structure must be compressed into a compact latent reasoning trajectory. In many instances, PLUME still places semantically related candidates near the top, indicating that the latent rollout preserves a substantial portion of the relevant query intent even when the exact target is not ranked first.

\clearpage

\clearpage

More broadly, the weakness on QA-style subsets should also be interpreted in light of the benchmark formulation itself. In MMEB, visual question answering is cast as an embedding retrieval problem, where the query consists of an instruction, an image, and a question, and the model must retrieve the correct answer from a fixed candidate pool. Prior work has noted that this type of evaluation pattern is not fully aligned with realistic question answering settings, where relevant evidence retrieval and answer generation are more tightly coupled, and where bounded candidate pools can simplify or distort the underlying retrieval--reasoning difficulty. We therefore view the larger gaps on text-rich QA benchmarks such as ChartQA and InfographicsVQA as reflecting both a model-side challenge in preserving fine-grained textual and numerical structure, and a benchmark-side mismatch between QA and retrieval formulations.

These cases are consistent with the discussion in Sec.~\ref{sec:analysis}: while latent reasoning is more efficient and often sufficient for retrieval-oriented representation formation, explicit verbal reasoning may still preserve certain types of fine-grained semantic structure better in some challenging settings.

\section{Additional Routing Visualization}
\label{app:additional_heatmap}

To further analyze the behavior of the semantic-anchor-guided routed adapter, we visualize the average activation rate of each expert across different input modalities. Rather than focusing on individual examples, this aggregated view provides a coarse but informative picture of how the routed adapter allocates computation under different modality conditions.

\begin{figure}[t]
    \centering
    \includegraphics[width=\linewidth]{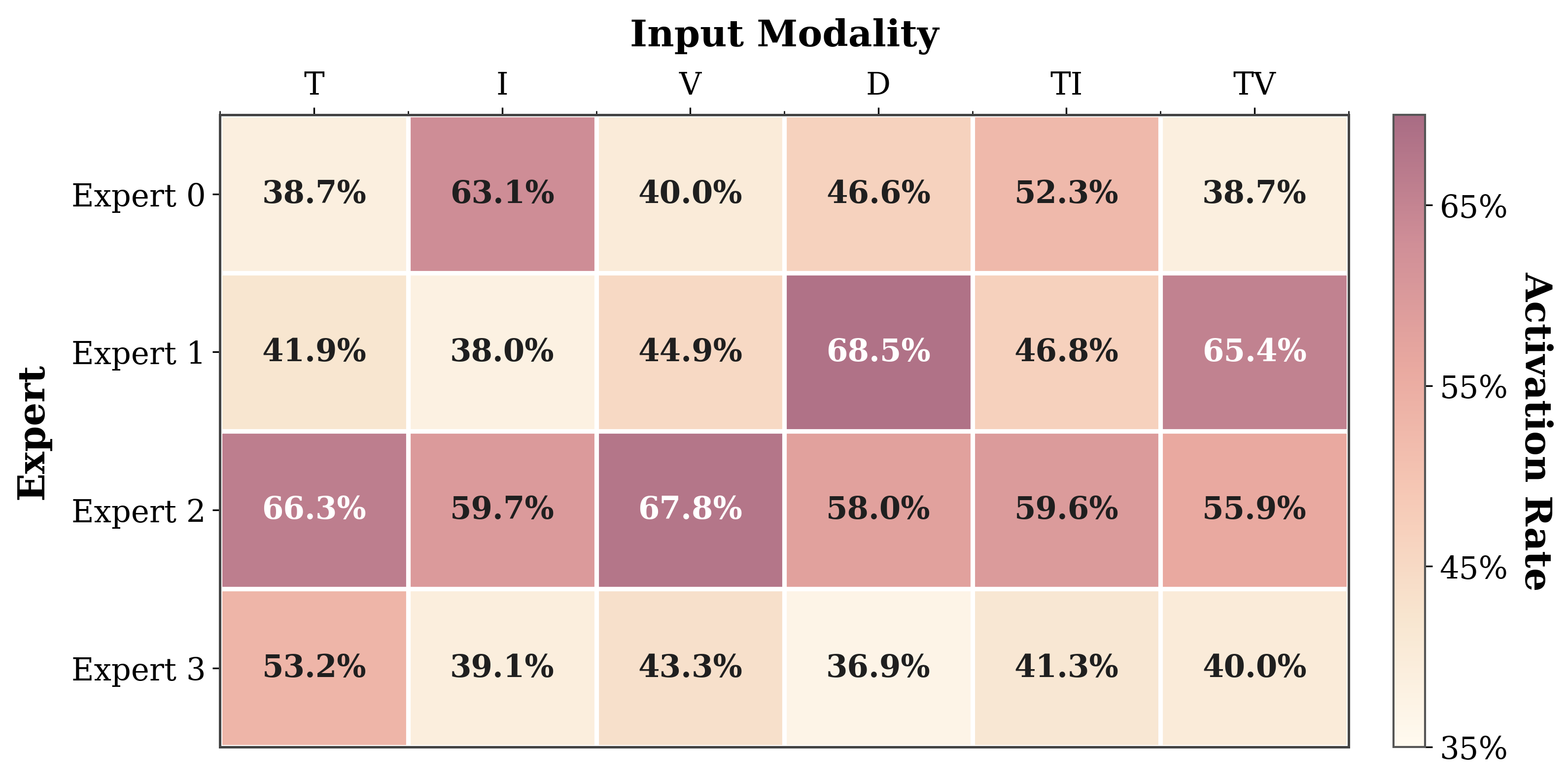}
    \caption{Average activation rate of each expert across input modalities. T, I, V, D, TI, and TV denote text, image, video, document, text-image, and text-video inputs, respectively. Darker cells indicate higher expert activation rates.}
    \label{fig:additional_routing_vis}
\end{figure}

Figure~\ref{fig:additional_routing_vis} summarizes expert activation patterns for six input types: text (T), image (I), video (V), document (D), text-image (TI), and text-video (TV). Several non-uniform trends can be observed. Expert 2 shows consistently high activation across almost all modalities, with especially strong responses on text and video inputs, suggesting that it serves as a broadly useful expert shared across diverse reasoning situations. In contrast, Expert 1 exhibits clearer specialization, reaching its highest activation on document and text-video inputs, which indicates a stronger preference for inputs with richer structured or cross-modal information. Expert 0 is comparatively more active on image and text-image inputs, while Expert 3 is most activated on text inputs and remains less preferred for document and mixed-modality cases.

These patterns suggest that the routed adapter does not collapse to a uniform expert usage distribution. Instead, different experts develop differentiated modality preferences, while still maintaining partial overlap that allows computation to be shared across related settings. This behavior is consistent with our design motivation: multimodal retrieval inputs are heterogeneous in structure, and thus benefit from adaptive latent transition pathways rather than a single fixed computation pattern.

Overall, the visualization shows that the routed adapter learns structured but non-uniform expert allocation patterns across modalities, supporting the claim that different multimodal inputs benefit from differentiated latent computation.

\section{Full MMEB-V2 Results}
\label{app:full_mmeb_results}

Tables~\ref{tab:full_mmeb_selected_image} and~\ref{tab:full_mmeb_selected_video_visdoc} give the per-dataset MMEB-v2 scores for PLUME-2B, UME-R1-2B, and selected baselines. Table~\ref{tab:main_results} reports the 7B comparison by sub-task and modality. These results show that the gains of PLUME are broadly distributed across task families, rather than arising from only a small subset of benchmarks.

\section{Additional Baseline Comparisons}
\label{app:additional_compare}
\begin{figure*}[t]
    \centering
    \includegraphics[width=\textwidth,height=0.89\textheight,keepaspectratio]{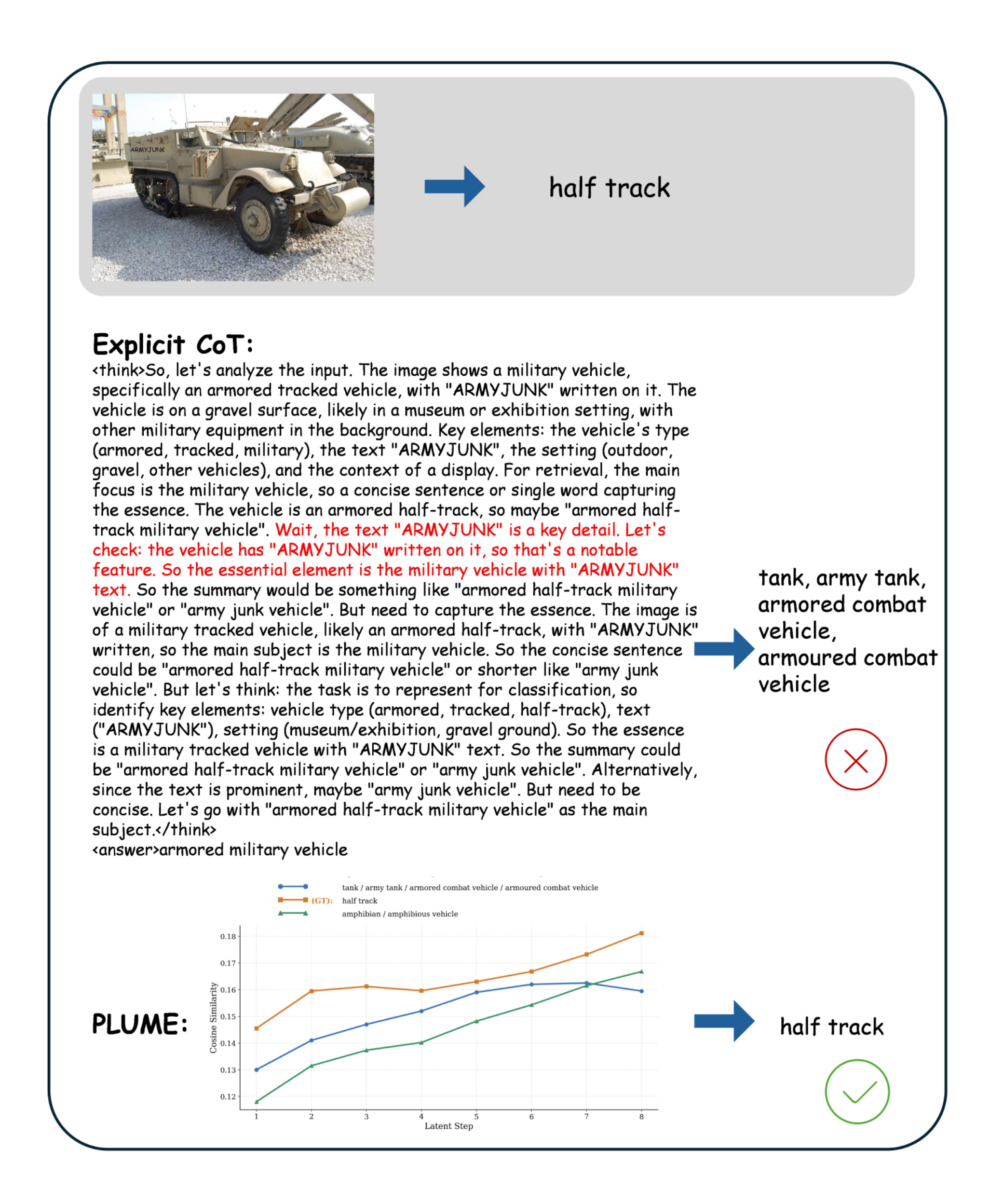}
    \caption{\textbf{A case where explicit CoT gets trapped by a misleading textual detail while PLUME recovers the correct target.}
    In this example, explicit CoT over-focuses on the spurious text cue ``ARMYJUNK,'' which biases the subsequent reasoning trace and leads to an incorrect retrieval result. In contrast, PLUME does not commit to a discrete verbal rationale and still retrieves the correct target \emph{half track} after latent rollout.}
    \label{fig:cot_vs_plume_recovery}
\end{figure*}

\begin{figure*}[!t]
    \centering
    \includegraphics[width=\textwidth,height=0.89\textheight,keepaspectratio]{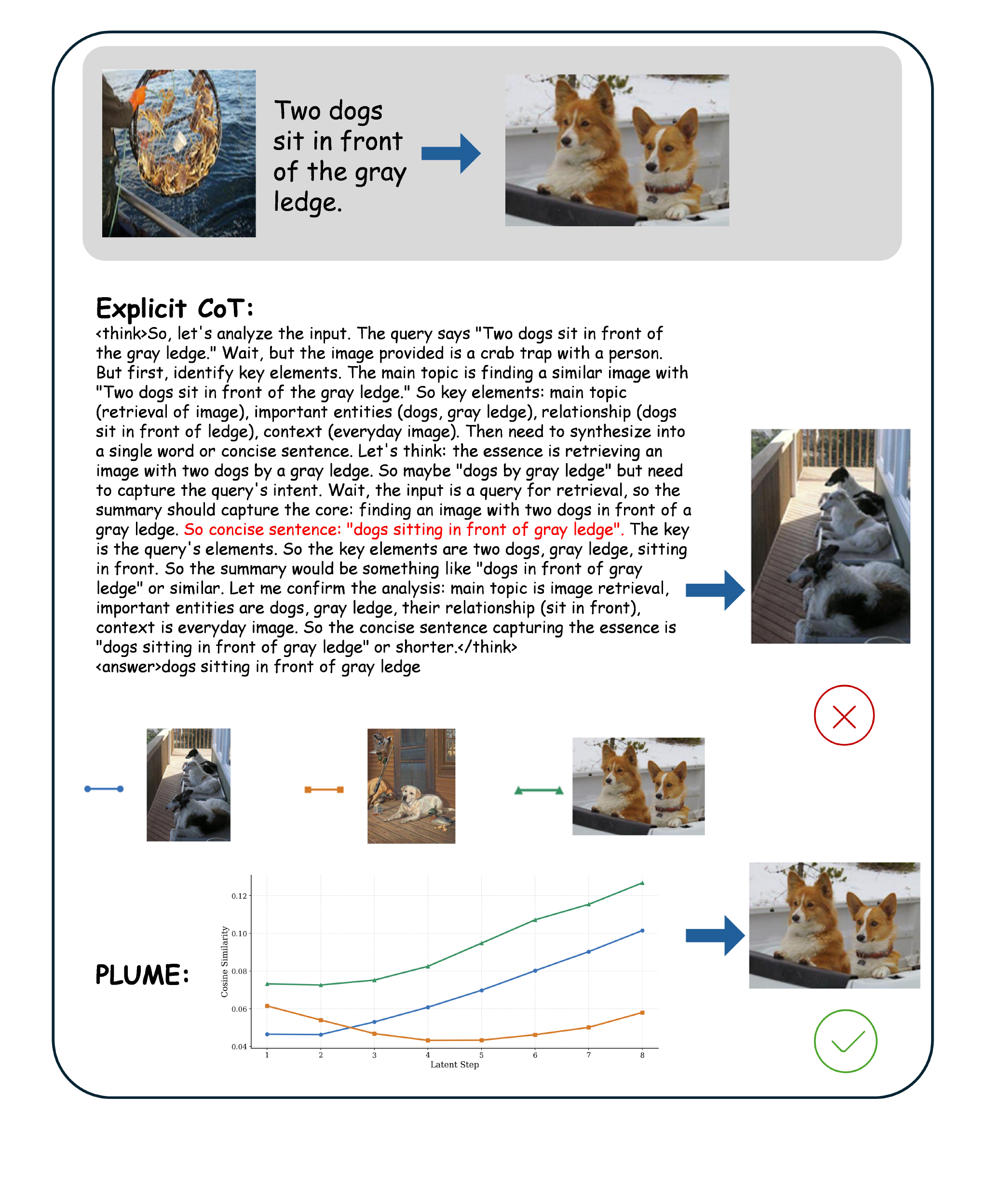}
    \caption{\textbf{A case where explicit CoT over-compresses the retrieval intent while PLUME preserves the correct target.}
    In this example, explicit CoT summarizes the query into an overly coarse verbal description and is distracted toward an incorrect retrieval result. In contrast, PLUME avoids committing to a discrete textual rationale and successfully retrieves the correct target after latent rollout.}
    \label{fig:cot_vs_plume_recovery_cirr}
\end{figure*}

\begin{figure*}[!t]
    \centering
    \includegraphics[width=\textwidth,height=0.89\textheight,keepaspectratio]{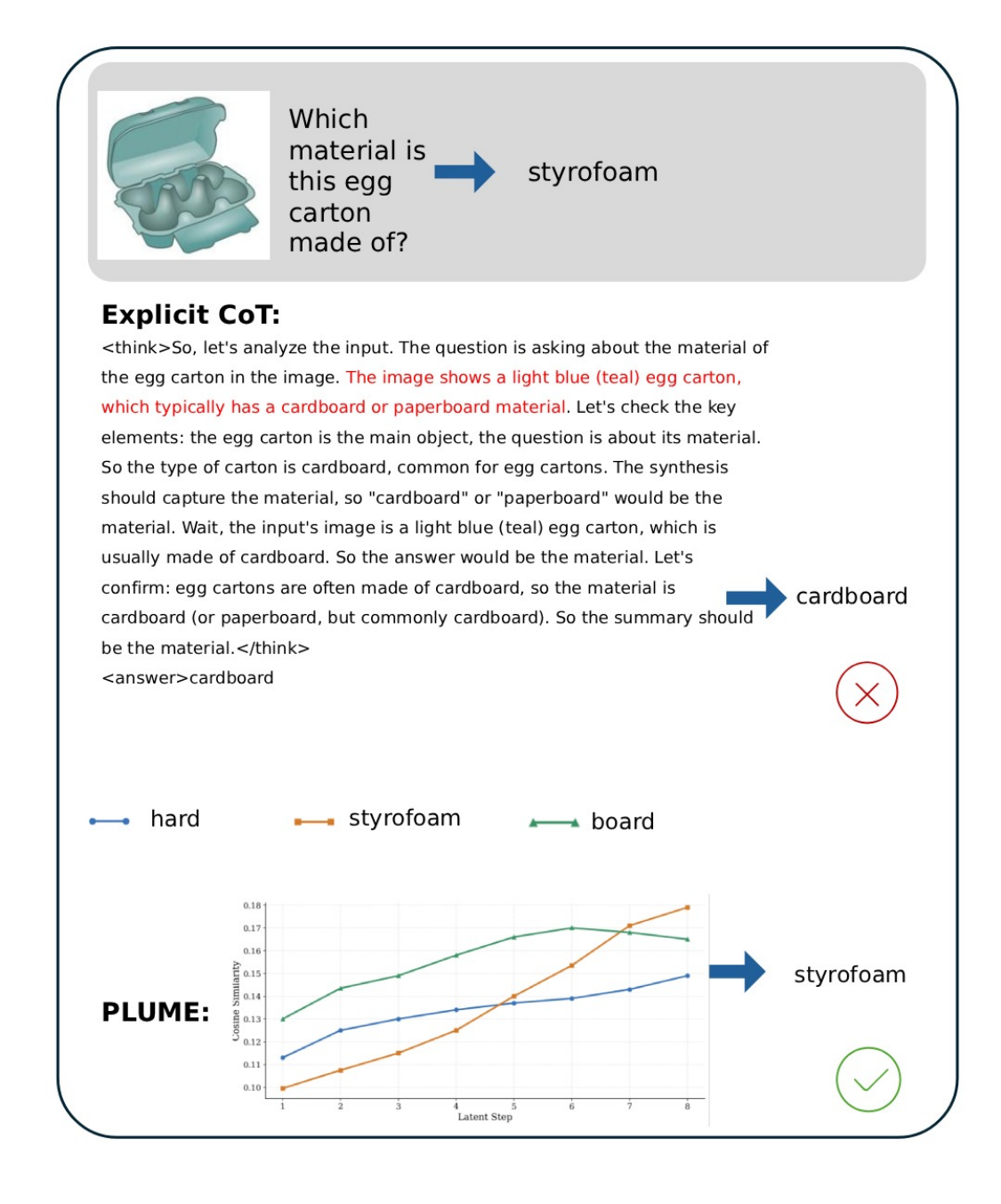}
    \caption{\textbf{A case where explicit CoT is misled by a surface-level prior while PLUME recovers the correct answer.}
    In this example, explicit CoT over-relies on the visual prior that egg cartons are typically made of cardboard and therefore predicts an incorrect result. In contrast, PLUME avoids being locked into this explicit verbal path and successfully retrieves the correct answer \emph{styrofoam} after latent rollout.}
    \label{fig:cot_vs_plume_recovery_scienceqa}
\end{figure*}

\begin{figure*}[!t]
    \centering
    \includegraphics[width=\textwidth,height=0.89\textheight,keepaspectratio]{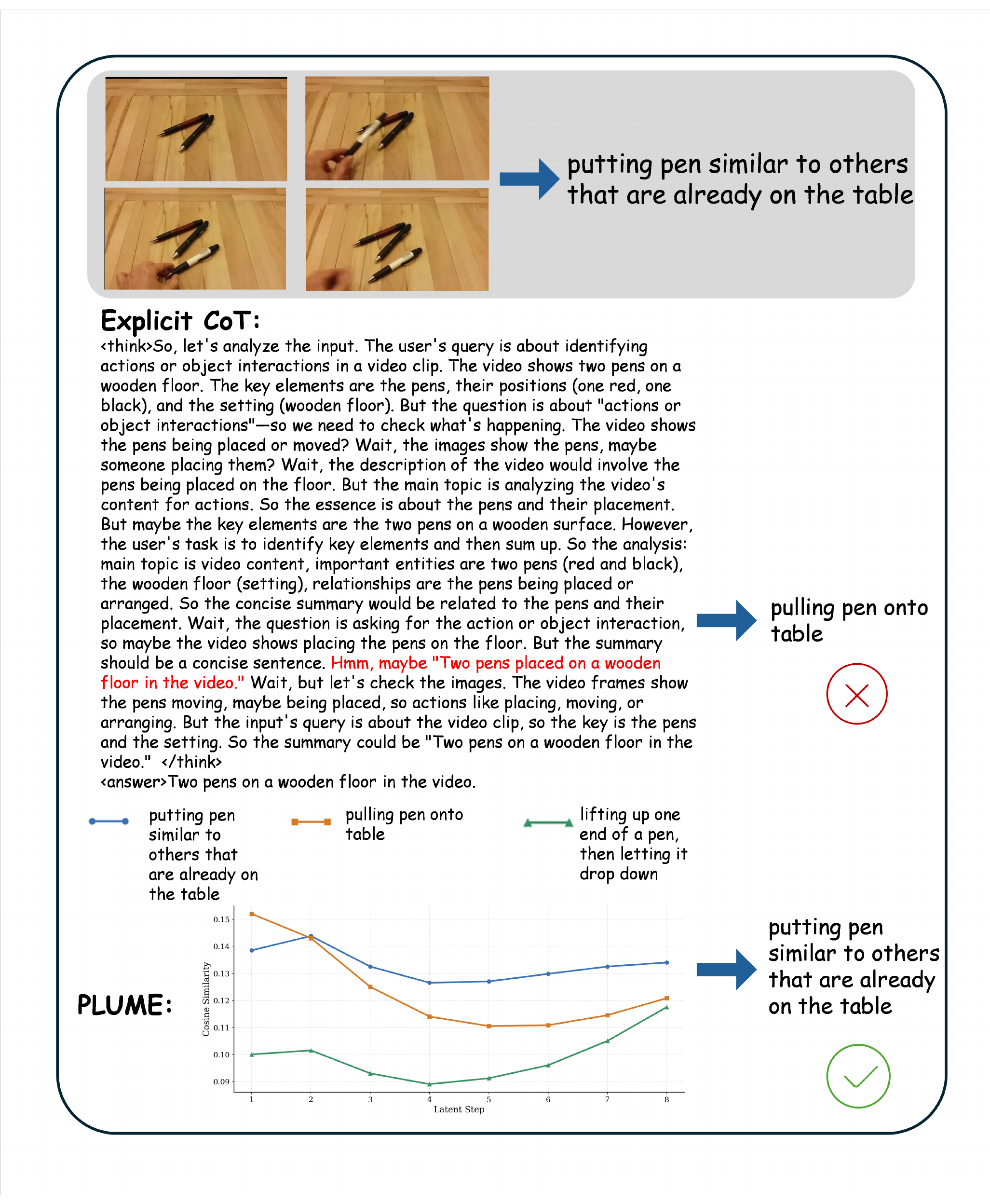}
    \caption{\textbf{A case where explicit CoT drifts toward an incorrect action description while PLUME recovers the correct target.}
    In this example, explicit CoT over-compresses the video into a coarse description of two pens on a wooden floor and is distracted toward an incorrect retrieval result. In contrast, PLUME avoids committing to this explicit verbal path and successfully retrieves the correct target \emph{putting pen similar to others that are already on the table} after latent rollout.}
    \label{fig:cot_vs_plume_recovery_smth}
\end{figure*}

In this section, we provide additional qualitative comparisons between explicit CoT and PLUME on representative retrieval cases. As shown in Figures~\ref{fig:cot_vs_plume_recovery}, \ref{fig:cot_vs_plume_recovery_cirr}, \ref{fig:cot_vs_plume_recovery_scienceqa}, and \ref{fig:cot_vs_plume_recovery_smth}, explicit CoT can fail in several different ways: it may over-focus on a spurious textual detail, compress the retrieval intent into an overly coarse verbal summary, rely on a misleading surface-level prior, or drift toward an incorrect action description. 

For each PLUME example, the small line chart visualizes the cosine similarity between the hidden state at each latent rollout step and the embeddings of the top retrieved candidates at the end of inference. In other words, the curves show how strongly each intermediate latent state aligns with several final candidate targets as the rollout proceeds. We emphasize that these trajectories are used only as a diagnostic view of latent computation: they are not intended to imply that every step must monotonically approach the correct target or correspond to an interpretable verbal reasoning trace. Instead, they illustrate a different property from explicit CoT: PLUME does not externalize intermediate reasoning into a fixed textual path, and therefore is less likely to be locked into an early verbal mistake. Across these cases, although the relative similarities may fluctuate during rollout, the correct target remains recoverable and is ultimately selected at the end. These examples further support our claim that latent reasoning is less vulnerable to discrete verbal lock-in than explicit CoT in retrieval-oriented embedding formation.

\begin{table*}[p]
\centering
\scriptsize
\setlength{\tabcolsep}{4.2pt}
\renewcommand{\arraystretch}{0.98}
\caption{Full MMEB-V2 results with selected baselines (Part I: Overall and Image). ``VLM2Vec'' denotes VLM2Vec-7B.}
\label{tab:full_mmeb_selected_image}
\resizebox{\textwidth}{!}{
\begin{tabular}{lccccccc}
\toprule
\textbf{Task} & \textbf{GME-2B} & \textbf{LamRA} & \textbf{VLM2Vec} & \textbf{VLM2Vec-V2.0} & \textbf{DUME-2B} & \textbf{UME-R1-2B} & \textbf{PLUME-2B} \\
\midrule
\rowcolor{blue!8}
Avg - All (78 tasks) & 54.1 & 40.5 & 46.9 & 58.1 & 52.7 & 60.1 & 61.6 \\
\rowcolor{blue!8}
Avg - Image (36 tasks, Hit@1) & 51.9 & 54.1 & 59.7 & 64.9 & 62.5 & 66.6 & 66.3 \\
\midrule
\rowcolor{orange!10}
I-CLS (10) & 54.6 & 59.3 & 58.6 & 62.9 & 59.3 & 64.8 & 66.5 \\
\rowcolor{orange!10}
I-QA (10) & 29.8 & 26.5 & 49.2 & 56.4 & 54.9 & 62.8 & 59.2 \\
\rowcolor{orange!10}
I-RET (12) & 66.8 & 70.1 & 65.0 & 69.6 & 66.3 & 67.6 & 67.6 \\
\rowcolor{orange!10}
I-VG (4) & 55.6 & 62.6 & 73.1 & 77.1 & 78.0 & 77.2 & 79.7 \\
\midrule
ImageNet-1K & 58.1 & 72.6 & 77.5 & 80.8 & 74.6 & 75.3 & 74.1 \\
N24News & 50.3 & 51.6 & 73.8 & 73.0 & 69.7 & 81.1 & 81.1 \\
HatefulMemes & 53.7 & 49.1 & 57.9 & 55.8 & 65.3 & 75.2 & 75.5 \\
VOC2007 & 75.8 & 80.0 & 74.3 & 84.9 & 68.9 & 80.0 & 86.1 \\
SUN397 & 67.0 & 68.7 & 73.7 & 70.9 & 71.4 & 79.4 & 76.9 \\
Place365 & 36.1 & 40.5 & 35.0 & 36.1 & 41.0 & 42.6 & 42.4 \\
ImageNet-A & 28.4 & 47.2 & 50.6 & 47.6 & 41.3 & 50.4 & 50.8 \\
ImageNet-R & 78.5 & 88.4 & 84.7 & 89.3 & 90.7 & 88.7 & 87.5 \\
ObjectNet & 71.2 & 66.3 & 36.9 & 65.1 & 46.2 & 52.0 & 61.5 \\
Country211 & 26.4 & 28.3 & 21.7 & 25.8 & 23.9 & 23.4 & 25.0 \\
OK-VQA & 30.1 & 37.9 & 48.2 & 51.7 & 56.8 & 62.4 & 60.5 \\
A-OKVQA & 18.6 & 26.9 & 39.7 & 44.0 & 46.9 & 51.1 & 49.9 \\
DocVQA & 30.0 & 22.4 & 82.7 & 90.1 & 86.0 & 92.2 & 89.9 \\
InfographicsVQA & 11.8 & 16.5 & 47.3 & 59.1 & 59.2 & 67.7 & 59.6 \\
ChartQA & 13.3 & 11.6 & 42.2 & 48.1 & 39.1 & 64.9 & 49.8 \\
Visual7W & 15.6 & 19.8 & 50.9 & 52.8 & 46.9 & 54.1 & 47.6 \\
ScienceQA & 27.1 & 26.5 & 30.5 & 38.1 & 38.7 & 42.7 & 42.9 \\
VizWiz & 37.1 & 31.9 & 38.8 & 43.3 & 42.0 & 46.8 & 46.5 \\
GQA & 75.3 & 38.3 & 48.1 & 65.4 & 60.2 & 67.3 & 69.1 \\
TextVQA & 39.5 & 33.1 & 63.2 & 71.6 & 73.9 & 78.6 & 78.9 \\
VisDial & 47.7 & 61.0 & 75.1 & 82.7 & 75.9 & 76.6 & 72.6 \\
CIRR & 43.1 & 52.1 & 46.8 & 57.3 & 52.0 & 53.7 & 54.6 \\
VisualNews.t2i & 74.8 & 70.9 & 73.4 & 74.7 & 71.2 & 71.7 & 71.3 \\
VisualNews.i2t & 77.7 & 84.1 & 73.8 & 78.3 & 72.5 & 74.2 & 72.7 \\
MSCOCO.t2i & 68.3 & 72.0 & 73.1 & 75.9 & 74.5 & 75.1 & 74.1 \\
MSCOCO.i2t & 63.2 & 73.6 & 68.3 & 71.1 & 68.3 & 68.9 & 69.8 \\
NIGHTS & 67.6 & 65.7 & 65.8 & 68.4 & 67.5 & 67.2 & 68.0 \\
WebQA & 88.8 & 81.2 & 85.8 & 90.6 & 90.2 & 90.0 & 89.1 \\
FashionIQ & 32.2 & 41.7 & 13.8 & 19.6 & 11.5 & 17.1 & 20.3 \\
Wiki-SS-NQ & 73.8 & 70.1 & 54.6 & 67.6 & 60.0 & 62.0 & 68.6 \\
OVEN & 72.1 & 82.2 & 68.4 & 64.8 & 65.2 & 66.9 & 68.4 \\
EDIS & 91.7 & 86.1 & 81.4 & 84.2 & 86.5 & 88.0 & 81.8 \\
MSCOCO & 28.4 & 44.7 & 65.7 & 66.2 & 68.1 & 69.5 & 66.9 \\
RefCOCO & 55.8 & 62.5 & 80.8 & 87.0 & 85.1 & 83.3 & 86.5 \\
RefCOCO-Matching & 73.7 & 76.2 & 76.6 & 86.3 & 89.3 & 84.4 & 88.4 \\
Visual7W-Pointing & 64.6 & 67.1 & 69.1 & 69.0 & 69.5 & 71.5 & 74.9 \\
\bottomrule
\end{tabular}
}
\end{table*}

\begin{table*}[p]
\centering
\scriptsize
\setlength{\tabcolsep}{4.2pt}
\renewcommand{\arraystretch}{0.98}
\caption{Full MMEB-V2 results with selected baselines (Part II: Video and VisDoc). ``VLM2Vec'' denotes VLM2Vec-7B.}
\label{tab:full_mmeb_selected_video_visdoc}
\resizebox{\textwidth}{!}{
\begin{tabular}{lccccccc}
\toprule
\textbf{Task} & \textbf{GME-2B} & \textbf{LamRA} & \textbf{VLM2Vec} & \textbf{VLM2Vec-V2.0} & \textbf{DUME-2B} & \textbf{UME-R1-2B} & \textbf{PLUME-2B} \\
\midrule
\rowcolor{blue!8}
Avg - Video (18 tasks, Hit@1) & 33.6 & 35.2 & 28.6 & 34.7 & 33.2 & 42.2 & 44.1 \\
\rowcolor{blue!8}
Avg - Visdoc (24 tasks, NDCG@5) & 72.7 & 23.9 & 41.6 & 65.4 & 52.8 & 63.9 & 67.5 \\
\midrule
\rowcolor{green!10}
V-CLS (5) & 34.8 & 39.4 & 33.3 & 39.2 & 37.7 & 44.3 & 45.0 \\
\rowcolor{green!10}
V-QA (5) & 41.8 & 42.7 & 30.7 & 34.7 & 46.6 & 51.0 & 52.3 \\
\rowcolor{green!10}
V-RET (5) & 25.4 & 24.5 & 20.5 & 28.4 & 17.1 & 32.9 & 33.5 \\
\rowcolor{green!10}
V-MR (3) & 31.8 & 33.6 & 30.7 & 37.6 & 30.0 & 39.7 & 46.7 \\
\midrule
K700 & 34.9 & 42.3 & 31.0 & 38.2 & 22.7 & 35.8 & 42.2 \\
SmthSmthV2 & 30.1 & 36.1 & 30.8 & 43.0 & 37.7 & 44.1 & 44.8 \\
HMDB51 & 43.0 & 40.4 & 34.0 & 40.2 & 53.4 & 54.4 & 51.2 \\
UCF101 & 52.1 & 60.7 & 57.6 & 60.0 & 55.7 & 67.2 & 66.5 \\
Breakfast & 13.6 & 17.6 & 12.9 & 14.8 & 18.9 & 20.1 & 20.1 \\
MVBench & 37.6 & 37.3 & 30.5 & 33.6 & 48.8 & 49.9 & 47.4 \\
Video-MME & 34.0 & 34.1 & 27.1 & 30.8 & 39.2 & 41.7 & 40.0 \\
NExTQA & 39.4 & 43.7 & 20.2 & 20.9 & 55.2 & 59.9 & 57.3 \\
EgoSchema & 40.6 & 44.8 & 25.6 & 35.0 & 23.2 & 45.4 & 47.8 \\
ActivityNetQA & 57.2 & 53.8 & 49.9 & 53.0 & 66.7 & 57.8 & 69.2 \\
DiDeMo & 21.5 & 25.0 & 19.1 & 30.0 & 16.9 & 32.4 & 32.7 \\
MSR-VTT & 27.0 & 22.6 & 25.6 & 27.8 & 16.2 & 34.3 & 36.2 \\
MSVD & 47.3 & 46.4 & 37.5 & 47.3 & 34.9 & 55.4 & 56.1 \\
VATEX & 23.1 & 19.1 & 15.7 & 26.2 & 11.1 & 29.9 & 28.2 \\
YouCook2 & 7.9 & 9.3 & 4.4 & 10.6 & 0.06 & 12.7 & 14.5 \\
QVHighlight & 44.0 & 53.9 & 43.7 & 49.7 & 40.3 & 57.5 & 57.1 \\
Charades-STA & 14.3 & 10.9 & 12.9 & 20.1 & 16.1 & 20.4 & 19.4 \\
MomentSeeker & 37.1 & 36.0 & 35.4 & 42.9 & 33.7 & 41.2 & 63.5 \\
\midrule
\rowcolor{purple!10}
VD-Vidore-V1 (10) & 87.1 & 33.8 & 20.6 & 74.4 & 67.6 & 72.4 & 72.1 \\
\rowcolor{purple!10}
VD-Vidore-V2 (4) & 53.9 & 11.5 & 13.2 & 44.6 & 43.3 & 46.2 & 49.8 \\
\rowcolor{purple!10}
VD-VisRAG (6) & 82.4 & 37.6 & 52.2 & 79.3 & 47.1 & 79.2 & 78.1 \\
\rowcolor{purple!10}
VD-OOD (4) & 43.1 & 21.0 & 33.6 & 39.4 & 33.8 & 37.2 & 57.4 \\
\midrule
ViDoRe.arxivqa & 82.5 & 31.5 & 18.1 & 78.9 & 68.7 & 73.9 & 72.6 \\
ViDoRe.docvqa & 55.2 & 19.9 & 14.0 & 37.1 & 33.6 & 37.9 & 36.2 \\
ViDoRe.infovqa & 90.7 & 63.7 & 39.5 & 82.7 & 74.5 & 76.2 & 79.0 \\
ViDoRe.tabfquad & 93.3 & 53.5 & 36.0 & 87.8 & 78.3 & 86.1 & 88.8 \\
ViDoRe.tatdqa & 70.3 & 7.9 & 10.5 & 44.3 & 35.3 & 40.6 & 36.6 \\
ViDoRe.shiftproject & 92.9 & 16.0 & 8.4 & 61.0 & 61.8 & 66.8 & 64.8 \\
ViDoRe.artificial\_intelligence & 98.1 & 29.8 & 17.0 & 89.1 & 74.3 & 85.9 & 83.8 \\
ViDoRe.energy & 92.6 & 36.1 & 16.4 & 86.3 & 78.4 & 83.3 & 82.6 \\
ViDoRe.government\_reports & 97.2 & 41.2 & 25.2 & 85.6 & 83.0 & 82.6 & 83.2 \\
ViDoRe.healthcare\_industry & 98.5 & 38.8 & 20.8 & 91.1 & 88.2 & 90.8 & 91.1 \\
ViDoRe.csg\_reports\_human\_labeled\_v2 & 60.3 & 6.9 & 13.1 & 45.8 & 48.0 & 50.2 & 52.1 \\
ViDoRe.biomedical\_lectures\_v2\_multilingual & 54.0 & 13.4 & 6.5 & 44.6 & 39.8 & 46.2 & 48.2 \\
ViDoRe.economics\_reports\_v2\_multilingual & 50.8 & 19.4 & 12.9 & 42.3 & 44.1 & 45.7 & 49.6 \\
ViDoRe.csg\_reports\_v2\_multilingual & 50.4 & 6.4 & 20.3 & 45.7 & 41.1 & 42.7 & 49.0 \\
VisRAG.ArxivQA & 82.0 & 2.0 & 41.2 & 76.7 & 35.8 & 74.3 & 71.6 \\
VisRAG.ChartQA & 79.1 & 42.7 & 59.5 & 84.2 & 47.2 & 86.0 & 80.8 \\
VisRAG.MP-DocVQA & 84.3 & 33.4 & 43.6 & 71.8 & 35.3 & 75.6 & 74.9 \\
VisRAG.SlideVQA & 93.7 & 56.3 & 74.5 & 91.4 & 61.3 & 87.1 & 88.9 \\
VisRAG.InfoVQA & 91.4 & 56.9 & 71.1 & 85.9 & 64.7 & 84.4 & 85.7 \\
VisRAG.PlotQA & 64.1 & 34.1 & 23.5 & 65.9 & 38.5 & 68.0 & 66.2 \\
ViDoSeck-page & 21.6 & 11.3 & 17.7 & 21.9 & 20.0 & 21.2 & 80.6 \\
ViDoSeck-doc & - & - & - & 79.0 & 69.5 & 75.9 & 76.9 \\
MMLongBench-page & 15.8 & 8.0 & 9.6 & 11.9 & 10.4 & 11.9 & 39.9 \\
MMLongBench-doc & - & - & - & 63.0 & 35.4 & 39.7 & 32.0 \\
\bottomrule
\end{tabular}
}
\end{table*}
\clearpage

\end{document}